%% file: main.tex
\PassOptionsToPackage{dvipsnames,table}{xcolor}
\documentclass[lettersize, journal]{IEEEtran}
\usepackage{amsmath,amsfonts}
\usepackage{algorithmic}
\usepackage{array}
\usepackage[font=normalsize,labelfont=sf,textfont=sf]{subfig}
\usepackage{textcomp}
\usepackage{stfloats}
\usepackage{url}
\usepackage{verbatim}
\usepackage{cite}
\usepackage{amssymb}
\usepackage{amsmath}
\usepackage{tikz}
\usepackage{xcolor}
\usepackage{tabularx}
\usepackage{graphicx}
\usepackage{booktabs}
\usepackage{multirow}
\usepackage{dsfont}
\usepackage{mathtools}
\usepackage{cuted}
\usepackage[margin=1in]{geometry}
\usepackage{pifont}
\usepackage{threeparttable}
\usepackage{adjustbox}
\usepackage{bm}
\usepackage{multicol}
\usepackage{hyperref}
\definecolor{forestgreen}{RGB}{34,139,34}   
\definecolor{firebrick}{RGB}{178,34,34}     
\definecolor{darkblue}{RGB}{128,177,211}
\definecolor{darkorange}{RGB}{255,127,0}
\newcommand{\cmark}{\textcolor{forestgreen}{\ding{51}}}
\newcommand{\xmark}{\textcolor{firebrick}{\ding{55}}}
\usetikzlibrary{arrows.meta, positioning, calc}

\usepackage{balance}
\newif\ifanonymous
\anonymousfalse 
\begin{document}
\ifanonymous
    \title{LEARN: \underline{L}earning \underline{E}nd-to-End \underline{A}erial \underline{R}esource-Constrained Multi-Robot \underline{N}avigation}
    \author{Anonymous Authors%
    \thanks{Anonymous authorship for review purposes.}}
\else

    \title{LEARN: \underline{L}earning \underline{E}nd-to-End \underline{A}erial \underline{R}esource-Constrained Multi-Robot \underline{N}avigation}
    \author{Darren Chiu*, Zhehui Huang*, Ruohai Ge, and Gaurav S. Sukhatme%
    \thanks{* Denotes equal contribution. \textit{\{chiudarr, zhehuihu\}@usc.edu}. All authors are with the University of Southern California.}
    
    }
\fi
\maketitle 
\begin{strip}
    \vspace{-7em}
    \begin{center}
    \includegraphics[width=1.0\textwidth]{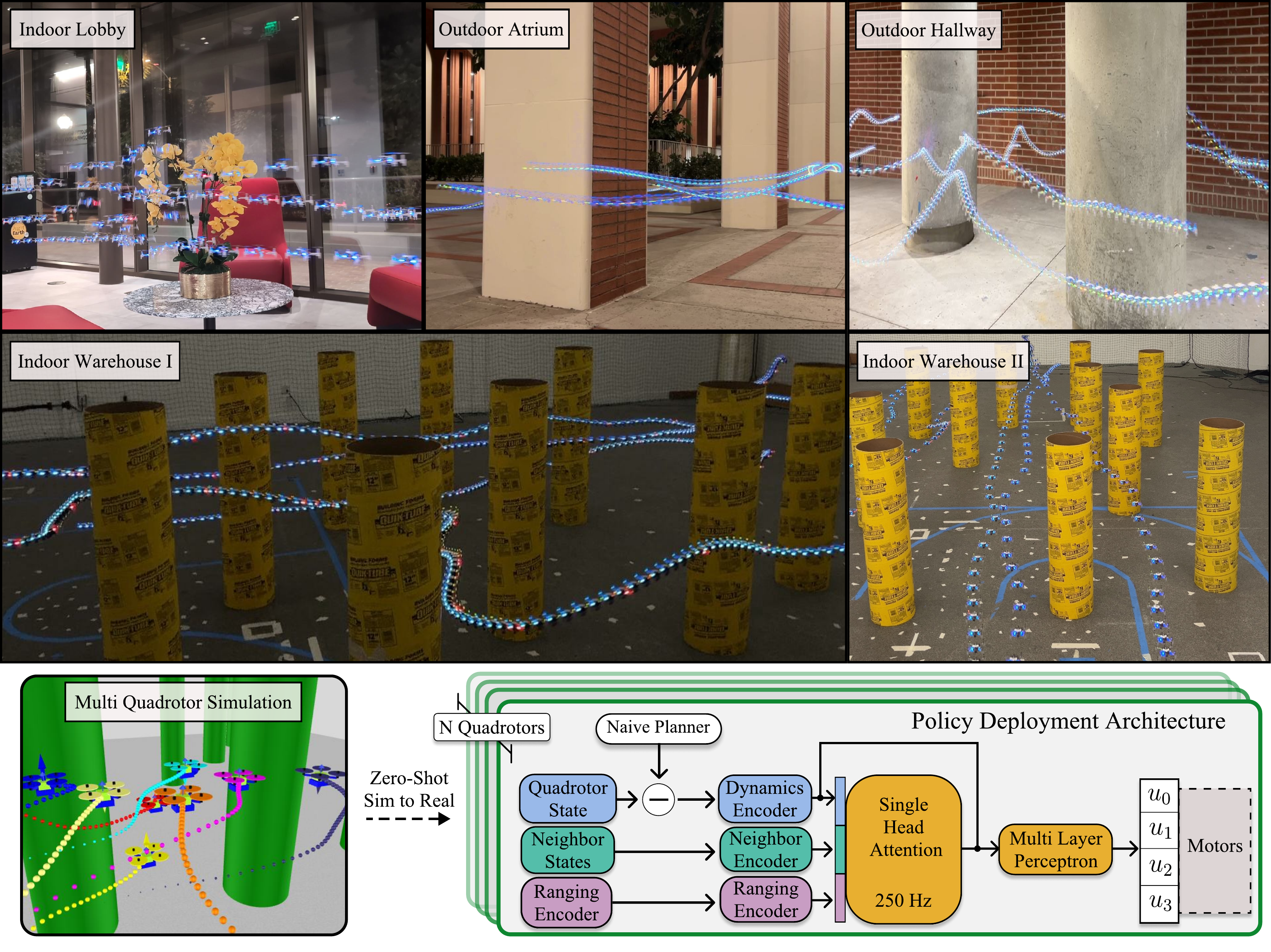}
    \captionof{figure}{LEARN is a lightweight, two-stage safety-guided reinforcement learning framework for multi-UAV navigation in cluttered indoor and outdoor spaces. All processes, including perception, localization, communication, planning, and control, run purely on an embedded single-core controller running at 168 MHz with 192 KB of RAM. A single policy is trained in simulation and duplicated across all quadrotors. During deployment, a minimum snap naive planner produces goal points for the encoder. Quadrotors obtain the two closest neighbor positions and velocities through radio; and obstacles are sensed using a low dimensional time-of-flight sensor. The policy generates individual normalized rotor thrusts that are sent directly to the motors. LEARN is zero-shot transferable to the real world with no fine-tuning. Experiments show that it scales up to 6 quadrotors in the real world and 24 in simulation. 
    }
    \label{fig:main_figure}
    \end{center}
    \vspace{-1em}
\end{strip}

\begin{abstract}

Nano-UAV teams offer great agility yet face severe navigation challenges due to constrained onboard sensing, communication, and computation. Existing approaches rely on high-resolution vision or compute-intensive planners, rendering them infeasible for these platforms. We introduce LEARN, a lightweight, two-stage safety-guided reinforcement learning (RL) framework for multi-UAV navigation in cluttered spaces. Our system combines low-resolution Time-of-Flight (ToF) sensors and a simple motion planner with a compact, attention-based RL policy. In simulation, LEARN outperforms two state-of-the-art planners by $10\%$ while using substantially fewer resources. We demonstrate LEARN's viability on six Crazyflie quadrotors, achieving fully onboard flight in diverse indoor and outdoor environments at speeds up to $2.0 m/s$ and traversing $0.2 m$ gaps.


\end{abstract}

\input{Sections/Introduction}

\input{Sections/RelatedWork}
\input{Sections/Background}
\input{Sections/problem_statement_language}
\begin{figure*}[hbtp]
\begin{center}
  \includegraphics[width=0.98\textwidth]{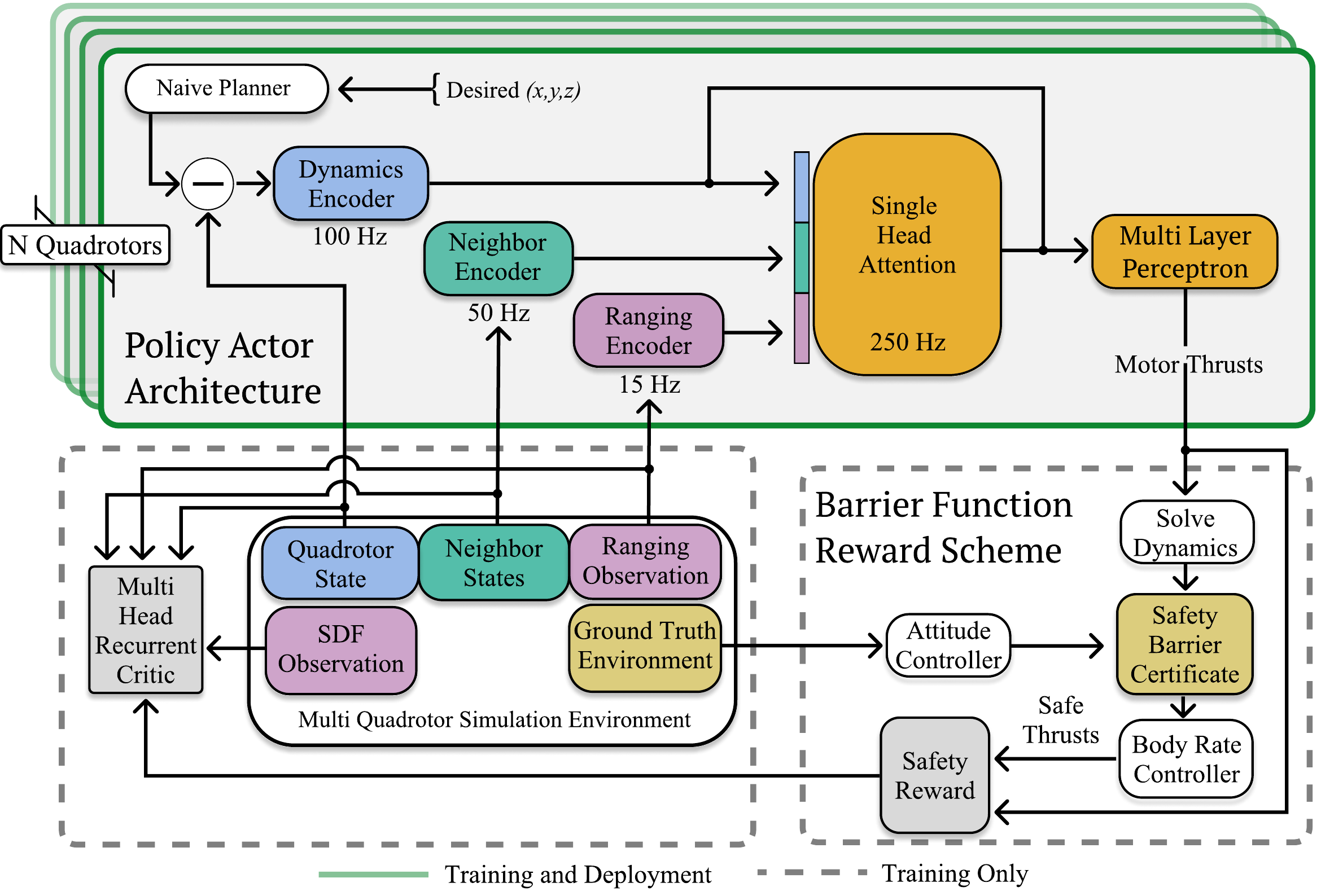}
  \caption{\textbf{Method Overview.} The training framework incorporates a two stage safety based reward function using Safety Barrier Certificates \cite{wang2017safety}. An asynchronous actor critic architecture is used where the critic observes a signed distance field (SDF) and employs a recurrent multi-headed attention architecture. At the beginning of each episode we generate a minimum snap trajectory. The trajectory is evaluated at each controller step to generate a 13 dimension goal point which is subtracted from the current state. The green denotes what is deployed on hardware and gray for purposes of training only. The same policy is both trained and deployed across all quadrotors.}
  \label{fig:system_overview}
  \vspace{-1em}
\end{center}
\end{figure*}
\input{Sections/Method}
\input{Sections/Results}
\input{Sections/Conclusion}
\bibliographystyle{IEEEtran}
\bibliography{./jb_ext}






\end{document}

%% file: Sections/Introduction.tex
\section{Introduction}
\IEEEPARstart{U}nmanned aerial vehicles (UAVs) are increasingly used in domains such as surveillance \cite{shakhatreh2019unmanned}, search and rescue \cite{mishra2020drone}, and planetary exploration \cite{hassanalian2018evolution}. 
The physics of flight impose stringent size, weight, and power (SWaP) constraints on these platforms, making efficient system design paramount.
While autonomy in UAVs has advanced significantly, many state-of-the-art navigation approaches fail to scale to resource-constrained platforms.
These approaches commonly deploy high-dimensional vision-based inputs using depth cameras \cite{zhou2022swarm} or LiDAR sensors \cite{arafat2023vision}. Furthermore, these methods assume access to ample computational resources and sufficient memory for map-building.
When scaling to multiple quadrotors, a high-bandwidth communication layer is usually needed to share state information. As such, sufficient computation, memory, and communication are key assumptions that many modern decentralized multi-quadrotor navigation approaches leverage \cite{toumieh2024hdsm, csenbacslar2024dream, honig2018trajectory, zhou2022swarm}. \looseness=-1

\input{Tables/related_work}

However, such assumptions do not hold for resource-constrained systems. In this work, we push these limits by imposing the following hardware requirements: \textit{(i)} single-core processors operating below 1 GHz clock speed; \textit{(ii)} onboard memory of less than 1 MB; and \textit{(iii)} wireless communication bandwidth below 10 MHz. As UAV platforms scale down in size, these constraints become more restrictive, precluding the use of high-power sensors and computationally demanding models. These limitations are further exacerbated in multi-robot systems, where real-time operation and decentralized decision-making must occur without reliance on a global map or reliable communication layer.
Deep reinforcement learning (DRL) has emerged as a promising alternative, offering the ability to learn complex behaviors that generalize beyond predefined maps. However, its application to resource-constrained robotics remains largely unexplored given assumptions for real-world deployment. \looseness=-1

We address two fundamental challenges that hinder the deployment of learning enabled resource-constrained robotic systems:

\paragraph{Learning End-to-end Control on Constrained Hardware}  
Modern deep learning models typically exceed the onboard memory capacity of nano-UAVs, preventing direct deployment. Additionally, the lack of parallel processing significantly slows inference, limiting control update rates and reducing flight stability. Prior work has shown that integrating classical planning and control techniques can improve robustness and agility in UAVs \cite{song2025actor}. Building on this insight, we introduce a hybrid framework that incorporates planning, safety-based control, and learning to achieve highly runtime efficient decentralized navigation while maintaining a minimal computational footprint.

\paragraph{Real World Generalization}  
Many existing solutions assume perfect knowledge of the world \cite{honig2018trajectory}, access to maps \cite{riviere2020glas}, or high-dimensional inputs \cite{zhou2022swarm}.
In contrast, we develop a trajectory-agnostic framework that uses only low-resolution time-of-flight depth sensors, assumes no global map, and is robust to noisy observations and communication delays. Our design prioritizes simplicity, robustness, and generalization; a minimal solution for real-time multi-quadrotor navigation under extreme hardware limitations.

We present an end-to-end learned system for a swarm of nano-UAVs that can track trajectories in densely cluttered environments while avoiding collisions with obstacles and neighboring quadrotors. Our approach is designed for challenging indoor and outdoor environments where no global map or centralized planner is available. The proposed system scales to 24 quadrotors in simulation and can successfully navigate through gaps merely $0.1m$ larger than the quadrotor body. Without retraining, it achieves real world trajectory tracking with flight speeds of up to $2.0 m/s$ and a mean positional error of less than 0.1 meters. Our main contributions are as follows: \looseness=-1
\begin{enumerate}
    \item We present LEARN, the first fully onboard, reinforcement learning-based navigation system for nano-UAV swarms in densely cluttered environments under stringent resource constraints. 
    \item We enhance robustness by integrating a two stage control barrier function reward into training and learn a trajectory aware policy with minimal sensing.
    \item  We evaluate LEARN extensively in simulation against two state-of-the-art methods, achieving $10\%$ higher success rates with dramatically lower compute and sensing usage. We show LEARN is zero-shot deployable in indoor and outdoor environments with fully onboard deployment using the Crazyflie 2.1.
\end{enumerate}

%% file: Tables/related_work.tex
\begin{table*}[t!]
  \centering
  \caption{Comparison of recent multi-robot collision avoidance and navigation methods}
  \label{table:related_works}
  \begin{threeparttable}
  \resizebox{\textwidth}{!}{%
\begin{tabular}{@{} l l c c c c c c c l c c @{}}
  \toprule
  Year 
  & Method 
  & Decentr. 
  & Async 
  & Realistic
  & \multicolumn{3}{c}{Onboard} 
  & \multicolumn{2}{c}{Resource‐constrained} 
  & Learning‐based 
  & Collision avoidance in \\
  \cmidrule(lr){6-8} \cmidrule(lr){9-10}
  &     &  
  &     & Observation   
  & Localization & Perception & Computation    
  & Yes/No?     & Platform         
  &             &                        \\
  \midrule
      \multirow{1}{*}{2018} & H\"onig et al.~\cite{honig2018trajectory}     & \xmark        & \xmark & \xmark
                  & \xmark         & \xmark         & \xmark        
                  & \cmark & Crazyflie        & \xmark        & Planning              \\
         \midrule

      \multirow{2}{*}{2019} & RSFC~\cite{park2020efficient}     & \xmark        & \xmark & \xmark
                  & \xmark         & \xmark         & \xmark        
                  & \cmark & Crazyflie        & \xmark        & Planning              \\
       & SGBA~\cite{mcguire_2019}     & \cmark        & \cmark & \cmark
                  & \cmark         & \cmark         & \cmark        
                  & \cmark & Crazyflie        & \xmark        & Control               \\
      \midrule
      \multirow{3}{*}{2020} & DMPC~\cite{luis2020online}     & \cmark        & \xmark & \xmark
                  & \xmark         & \xmark         & \xmark        
                  & \cmark & Crazyflie        & \xmark        & Planning              \\
      & Dec-RSFC~\cite{park2020online}      & \cmark        & \xmark & \xmark
                  & \xmark         & \xmark         & \xmark        
                  & \cmark & Crazyflie        & \xmark        & Planning              \\
      & GLAS~\cite{riviere2020glas}     & \cmark        & \cmark & \xmark
                  & \xmark         & \xmark         & \cmark        
                  & \cmark & Crazyflie        & \cmark        & Control               \\
      \midrule
      \multirow{4}{*}{2021} & PCAS~\cite{soria2021predictive}     & \xmark        & \xmark & \xmark 
                  & \xmark         & \xmark         & \xmark        
                  & \cmark & Crazyflie        & \xmark        & Control                   \\
      & DPDS~\cite{soria2021distributed}     & \cmark        & \xmark & \xmark
                  & \xmark         & \xmark         & \xmark        
                  & \cmark & Crazyflie        & \xmark        & Control                   \\
      & Zhu et al.~\cite{zhu2021learning}  & \cmark        & \cmark & \xmark
                  & \xmark         & \xmark         & \xmark        
                  & \xmark & Parrot Bebop 2   & \cmark        & Control              \\
      & EGO‐Swarm~\cite{zhou2021ego} & \cmark        & \cmark & \cmark 
                  & \cmark         & \cmark         & \cmark        
                  & \xmark & NVIDIA Xavier NX\tnote{*} & \xmark        & Planning              \\
      \midrule
      \multirow{4}{*}{2022} & EGO‐Swarm2~\cite{zhou2022swarm}
                  & \cmark        & \cmark & \cmark
                  & \cmark         & \cmark         & \cmark        
                  & \xmark & NVIDIA Xavier NX\tnote{*} & \xmark        & Planning              \\
      & EDG-Team~\cite{hou2022enhanced} & \cmark$|$\xmark\tnote{1} & \cmark$|$\xmark\tnote{1} & \cmark
                  & \cmark         & \cmark         & \cmark        
                  & \xmark & NVIDIA Xavier NX\tnote{*} & \xmark        & Planning              \\
      & LSC~\cite{park2022online}   & \cmark        & \xmark & \xmark
                  & \xmark         & \xmark         & \xmark        
                  & \cmark & Crazyflie        & \xmark        & Planning              \\
      & Ryou et al.~\cite{ryou2022cooperative}   & \xmark        & \xmark & \xmark
                  & \xmark         & \xmark         & \xmark        
                  & \xmark & NVIDIA Jetson TX2\tnote{*}        & \xmark        & Planning              \\
      \midrule
      \multirow{6}{*}{2023} & LSC‐DR~\cite{park2023decentralized}   & \cmark        & \xmark & \xmark
                  & \xmark         & \xmark         & \xmark        
                  & \cmark & Crazyflie        & \xmark        & Planning              \\
      & DLSC~\cite{park2023dlsc}     & \cmark        & \xmark & \xmark
                  & \xmark         & \xmark         & \xmark        
                  & \cmark & Crazyflie        & \xmark        & Planning              \\
      & RLSS~\cite{csenbacslar2023rlss}     & \cmark        & \xmark & \xmark
                  & \xmark         & \xmark         & \xmark        
                  & \cmark & Crazyflie        & \xmark        & Planning              \\
      & RMADER~\cite{kondo2023robust}   & \cmark        & \cmark & \xmark
                  & \xmark         & \xmark         & \cmark        
                  & \xmark & Intel NUC10\tnote{*}   & \xmark        & Planning              \\
      & IMPC‐OB~\cite{chen2023multi}  & \cmark        & \xmark & \xmark 
                  & \xmark         & \xmark         & \xmark        
                  & \cmark & Crazyflie        & \xmark        & Planning              \\
      & AMSwarm~\cite{adajania2023amswarm}  & \cmark        & \xmark & \xmark
                  & \xmark         & \xmark         & \xmark        
                  & \cmark & Crazyflie        & \xmark        & Planning              \\
      \midrule
      \multirow{8}{*}{2024} & AMSwarmX~\cite{adajania2024amswarmx} & \cmark        & \xmark & \xmark
                  & \xmark         & \xmark         & \xmark        
                  & \cmark & Crazyflie        & \xmark        & Planning              \\
      & Safaoui et al.~\cite{safaoui2024safe} & \xmark        & \xmark & \xmark
                  & \xmark         & \xmark         & \xmark        
                  & \cmark & Crazyflie        & \xmark        & Control              \\ 
      & SOGM~\cite{wu2024decentralized}     & \cmark        & \cmark & \xmark
                  & \xmark         & \xmark         & \cmark        
                  & \xmark & NVIDIA Xavier NX\tnote{*} & \xmark        & Planning              \\
      & DREAM~\cite{csenbacslar2024dream}    & \cmark        & \cmark & \xmark
                  & \xmark         & \xmark         & \xmark        
                  & \cmark & Crazyflie        & \xmark        & Planning              \\
      & HDSM~\cite{toumieh2024hdsm}     & \cmark        & \xmark & \cmark
                  & \xmark         & \xmark         & \xmark        
                  & \cmark & Crazyflie        & \xmark        & Planning              \\
      & CAN~\cite{huang2024collision}   & \cmark        & \cmark & \cmark
                  & \xmark         & \xmark         & \cmark        
                  & \cmark & Crazyflie        & \cmark        & Control               \\
      & Dec‐NMPC~\cite{goarin2024decentralized} & \cmark        & \cmark & \xmark
                  & \xmark         & \xmark         & \cmark        
                  & \xmark & Qualcomm VOXL2\tnote{*}    & \xmark        & Control               \\
      \midrule
      \multirow{2}{*}{2025} & 
      Pan et al.~\cite{pan2025hierarchical}     & \xmark        & \xmark & \xmark
                  & \xmark         & \xmark         & \xmark        
                  & \cmark & Crazyflie        & \xmark        & Planning              \\ 
      & \textbf{LEARN (Ours)}
                  & \cmark        & \cmark & \cmark 
                  & \cmark         & \cmark         & \cmark        
                  & \cmark & Crazyflie        & \cmark        & Control               \\
      \bottomrule
    \end{tabular}%
  }
    \begin{tablenotes}[flushleft]
      \footnotesize
    \item[1] EDG-Team switches to a centralized and synchronous planner in dense environments~\cite{toumieh2024hdsm}.
    \item[*] The platform is customized.
    \end{tablenotes}
  \end{threeparttable}
  \vspace{-1em}
\end{table*}

%% file: Sections/RelatedWork.tex
\tikzset{
  paper/.style={
    draw,
    rectangle,
    rounded corners=5pt,
    align=center,
    minimum width=1.6cm,
    minimum height=0.6cm,
    font=\scriptsize
  },
  control/.style={paper, fill=forestgreen!50},
  planning/.style={paper, fill=darkorange!50},
  arrow/.style={
    ->,
    >=Stealth,
    thick,
    font=\scriptsize     
  }
}
\input{Tables/digraph}
\section{Related Work}
\label{sec:related_works}
In this section, we discuss existing works in three key areas: multi-robot collision avoidance and navigation, learning-based safe control, and fully onboard methods for resource-constrained robotic platforms. 
Table~\ref{table:related_works} presents a curated list of multi-robot navigation and collision-avoidance approaches that have been demonstrated in real-world settings. 
From this collection, we identify those methods that employ realistic obstacle perception and adopt them as our baselines. 
Figure~\ref{fig:citation_graph} presents a directed graph capturing the performance relationships among these baselines as well as the additional works they compare against.

\subsection{Multi-robot Collision Avoidance and Navigation}
Existing multi-robot collision avoidance and navigation methods often incorporate simplifying assumptions, limiting their applicability for practical deployments, particularly on resource-constrained robotic platforms. Key restrictive assumptions prevalent in the literature include: \looseness=-1
\begin{enumerate}
\item Ideal localization: neglect localization error and usually rely on external motion capture systems during deployment.
\item Unrealistic obstacle observations: assume \textit{a priori} known environments with reliance on external motion capture systems for guidance. 
\item Ideal communication: assumes no communication delays and usage of external motion capture systems to relay information. 
\item Computation-rich hardware: assume computation resources are ample and usually rely on powerful onboard hardware or ground station during deployment. \looseness=-1
\end{enumerate}
Many multi-robot collision-avoidance and navigation methods are designed assuming abundant onboard resources. When deployed on platforms with limited sensing, computation, or communication capabilities, these methods typically rely on external systems, such as motion-capture setups or ground stations, to compensate. These assumptions significantly hinder the feasibility of deployment on fully onboard, resource-constrained robots. \looseness=-1

The first family of methods~\cite{honig2018trajectory, luis2020online, soria2021distributed, zhu2021learning, chen2023multi, adajania2023amswarm, adajania2024amswarmx, park2023decentralized, park2023dlsc, kondo2023robust, csenbacslar2023rlss, csenbacslar2024dream, safaoui2024safe, toumieh2024hdsm, pan2025hierarchical} are comprehensively based on external systems for precise location, obstacle tracking or pre-computed maps, and powerful hardware on board or ground stations for computation and communication. 
Although most approaches assume ideal communication, some explicitly incorporate realistic conditions such as communication delay~\cite{kondo2023robust, csenbacslar2024dream, toumieh2024hdsm}. 
In addition, although HDSM~\cite{toumieh2024hdsm} relies on motion-capture systems for obstacle data during deployment, they present simulation results using a $360^\circ$ LIDAR.

The second family of methods~\cite{park2020efficient, soria2021predictive, ryou2022cooperative, goarin2024decentralized, wu2024decentralized} still relies on external systems for high‐precision localization, obstacle tracking, and precomputed maps, as well as offboard computational resources, but removes dependence on ground stations for communication.
Within this group, some approaches eliminate runtime communication by precomputing trajectories for known environments~\cite{honig2018trajectory, park2020efficient, soria2021predictive, ryou2022cooperative}, while others support inter-robot state exchange over wireless networks~\cite{goarin2024decentralized, wu2024decentralized}.

The third family of methods~\cite{riviere2020glas, huang2024collision} similarly relies on external motion-capture systems for robot and obstacle localization and on ground stations for communication, yet achieves computational efficiency suitable for resource-constrained deployment. GLAS~\cite{riviere2020glas} employs deep imitation learning with a decentralized policy comprising five multilayer perceptron (MLP) layers trained from a global planner and augmented by a safety filter to enhance collision avoidance; it runs at 40 Hz on the Crazyflie. ~\cite{huang2024collision} introduce an end-to-end deep reinforcement learning approach using a compact, attention-based model that maps raw observations directly to rotor thrust commands, achieving execution rates of $1kHz$ on the Crazyflie.

The fourth family of methods~\cite{zhou2021ego, zhou2022swarm, hou2022enhanced} eliminates reliance on external systems by using onboard sensors for robot and obstacle localization and onboard hardware for computation. Nonetheless, these approaches still require high-performance processors, high-bandwidth peer-to-peer communication networks, and sufficient memory for map storage. For example, Ego-Swarm~\cite{zhou2021ego}, Ego-Swarm2~\cite{zhou2022swarm}, and EDG-Team~\cite{hou2022enhanced} leverage stereo grayscale and depth images from an Intel RealSense D430, running on an NVIDIA Xavier NX (six-core CPU, 384-core GPU, 8 GB RAM) for real-time mapping and collision avoidance.

The fifth family, exemplified by SGBA~\cite{mcguire_2019}, removes the need for powerful onboard hardware, making it viable for fully resource-constrained platforms. SGBA uses a single VL53L1x ToF sensor and grayscale camera for localization, four additional VL53L1x sensors for omnidirectional obstacle detection, a Nordic nRF51822 radio (2 Mbps), and a 168 MHz single-core CPU with 192 KB of RAM. However, its narrow $27^\circ$ field of view and simple rule-based controller are ill-suited for densely cluttered settings and goal achieving. 

Inspired by SGBA’s minimal-resource design~\cite{mcguire_2019} and the high-frequency execution of Huang et al.~\cite{huang2024collision}, we propose an end-to-end, decentralized deep reinforcement learning framework that empowers multiple nano-scale robots to navigate robustly in densely cluttered, unstructured environments using only onboard sensing, computation, and communication.

\subsection{Learning With Safe Control}
Traditional control theory relies on an explicit dynamics model to design controllers that guarantee stability and safety under known operating conditions. In contrast, reinforcement learning follows a data-driven paradigm, trading formal guarantees for adaptability to complex, uncertain environments~\cite{brunke2022safe}. Hybrid “safe‐learning” approaches aim to combine these paradigms: they enforce provable safety constraints via model-based controllers or safety filters, while using learned policies to optimize performance and handle unmodeled dynamics. \looseness=-1

We categorize existing safe‐learning methods into four families, based on how safety is encoded:

The first family of methods uses learning‐based policies to generate control actions and then applies existing safety filters (i.e., control barrier functions, or CBFs) to enforce theoretical guarantees~\cite{riviere2020glas, cheng2019end, cui2022learning, xiao2023barriernet}. 
However, existing non‐adaptive CBFs are often overly conservative~\cite{csomay2021episodic, xiao2023barriernet}. 
Hence, the second family of methods learns the safety filter itself to reduce conservatism~\cite{gao2023online, xiao2023barriernet, so2024train}. 
Both the first and second families, however, require the safety filter to run at a high enough frequency to maintain their guarantees, contributing to the computational overhead. Moreover, they typically assume full knowledge of obstacle positions and shapes, which is an unrealistic requirement for onboard‐only deployment. 
To address this, the third family of methods abandons explicit safety filters and instead trains a value network to predict state safety~\cite{he2024agile}. If the network determines the current state is safe, the original policy proceeds. Otherwise, a backup policy is activated.
The fourth family retains the concept of safety filters but uses them only during training to shape the reward, rather than to correct actions at runtime~\cite{krasowski2022provably, wang2022ensuring, bejarano2024safety}. 
Krasowski et al.~\cite{krasowski2022provably} uses a constant penalty when the safety filter is activated. 
Wang et al.~\cite{wang2022ensuring} and Bejarano et al.~\cite{bejarano2024safety} scale the penalty by the magnitude of the proposed correction.

Our framework belongs to the fourth family. Unlike prior work, which assumes a single unified reward function, we draw inspiration from the third family to design a two-stage reward:  
\begin{itemize}
  \item \emph{Nominal stage:} If no safety filter is triggered, the reward balances goal progression, collision avoidance, stability, energy efficiency, and smoothness.  
  \item \emph{Safety stage:} Once the filter triggers, we mask energy and smoothness rewards to encourage maneuvers that restore safety.
\end{itemize}

\subsection{Resource Constrained Robots}
Despite significant advances in single and multi-quadrotor navigation regardless of planning, learning, and safety-based methods, deploying these techniques on severely resource-constrained nano platforms in cluttered environments remains an open challenge. Resource-constrained robots, limited by sensing range, battery capacity, computational power, and communication bandwidth, struggle to achieve reliable autonomy. A few studies have demonstrated fully onboard solutions for tasks such as source seeking~\cite{Duisterhof_2021_sniffy} and rudimentary navigation~\cite{mcguire_2019}, but they typically assume sparsely featured or uncluttered settings.

To the best of our knowledge, our work is the first to address fully onboard perception and control for robust navigation in densely cluttered, unstructured environments using severely resource-constrained robots. We present a lightweight deep reinforcement learning framework that seamlessly integrates safety-guided learning, decentralized multi-quadrotor coordination, and onboard perception, enabling efficient and robust autonomy on nano-scale platforms.

%% file: Tables/digraph.tex
\begin{figure}[t!]
  \centering
  \scriptsize
  \resizebox{\columnwidth}{!}{%
    \begin{tikzpicture}[
        node distance=0.4cm and 0.4cm
      ]

    \node[control] (ORCA) {ORCA~\cite{van2011reciprocal}};
    
    \node[control, below=0.8cm of ORCA]   (DPDS)  {DPDS~\cite{soria2021distributed}};
    \node[control, below left=0.8cm and 1.3cm of ORCA]    (BVC)   {BVC~\cite{zhou2017fast}};
    \node[planning, below right=0.8cm and 0.8cm of ORCA]   (DMPC)  {DMPC~\cite{luis2020online}};
    
    \coordinate (mid_r1) at ($(BVC)!0.5!(DMPC)$);
    \node[control, below left=0.8cm and 2.8cm of mid_r1]   (GLAS)  {GLAS~\cite{riviere2020glas}};
    
    \node[planning, below left=1.8cm and 0.1cm of mid_r1]   (LSC)  {LSC~\cite{park2022online}};
    \node[planning, left=0.4cm of LSC]   (AMSwarm)  {AMSwarm~\cite{adajania2023amswarm}};
    
    \node[planning, below right=1.8cm and 0.3cm of mid_r1]   (EDG-Team) {EDG-Team~\cite{hou2022enhanced}};
    \node[planning, right=1.0cm of EDG-Team]   (EGO-Swarm) {EGO-Swarm~\cite{zhou2021ego}};
    \node[planning, below right=4.3cm and 3.9cm of mid_r1]   (SOGM)  {SOGM~\cite{wu2024decentralized}};
    \node[planning, below right=0.6cm and 2.8cm of mid_r1]   (RSFC)  {RSFC~\cite{park2020efficient}};
    
    \node[control, below left=3.1cm and 2.8cm of mid_r1]       (SBC)       {SBC~\cite{wang2017safety}};
    
    \node[planning, below left=4.2cm and 0.2cm of mid_r1] (AMSwarmX)  {AMSwarmX~\cite{adajania2024amswarmx}};
    \node[planning,right=0.6cm of AMSwarmX]      (RMADER)   {RMADER~\cite{kondo2023robust}};
    \node[planning, below right=3.0cm and 3.2cm of mid_r1]        (MADER)    {MADER~\cite{tordesillas2021mader}};
    
    \node[planning, below=5.6cm of mid_r1] (HDSM) {HDSM~\cite{toumieh2024hdsm}};
    \node[control, left=1.8cm of HDSM] (CAN-RL) {CAN~\cite{huang2024collision}};
    \node[planning, right=2.1cm of HDSM] (EGOSwarm2) {EGO‐Swarm2~\cite{zhou2022swarm}};
    
    \node[control, below=7.2cm of mid_r1]    (Ours)      {LEARN};
    
    \draw[arrow] (CAN-RL)   -- (SBC) node[near start, above] { \cite{huang2024collision}};
    \draw[arrow] (CAN-RL.west)   to[out=120,in=240] node[midway, below] { \cite{huang2024collision}} (GLAS.west);
    \draw[arrow] (SBC)      -- (CAN-RL);
    
    \draw[arrow] (SBC.west) to[out=120,in=220]  node[near start, above]  { \cite{huang2024collision}} (GLAS);

    \draw[arrow] (GLAS.west) to[out=110,in=180]  node[midway, above] { \cite{riviere2020glas}} (ORCA.west);

    \draw[arrow] (HDSM)      -- (AMSwarmX)   node[near start, above] { \cite{toumieh2024hdsm}};
    \draw[arrow] (HDSM)      -- (RMADER)     node[near start, above] { \cite{toumieh2024hdsm}};
    \draw[arrow] (HDSM)      -- (MADER)      node[near start, above] { \cite{toumieh2024hdsm}};
    \draw[arrow] (HDSM)      -- (EGOSwarm2)  node[midway, above] { \cite{toumieh2024hdsm}};

    \draw[arrow] (EGOSwarm2.east) to[out=45,in=355] node[midway, below] { \cite{zhou2022swarm}} (EGO-Swarm.east);
    \draw[arrow] (EGOSwarm2) -- (HDSM);
    \draw[arrow] (EGOSwarm2.west) to[out=100,in=230] node[near start, above] { \cite{zhou2022swarm}} (MADER);

    \draw[arrow] (RMADER)    -- (MADER)      node[midway, above] { \cite{kondo2023robust}};
    \draw[arrow] (RMADER)    -- (EDG-Team)      node[near start, above] { \cite{kondo2023robust}};
    \draw[arrow] (RMADER)    -- (EGO-Swarm)      node[near start, above] { \cite{kondo2023robust}};

    \draw[arrow] (MADER.east) to[out=20,in=350] node[very near start, above] { \cite{tordesillas2021mader}} (RSFC);

    \draw[arrow] (SOGM) -- (MADER) node[near start, above] { \cite{wu2024decentralized}};
    \draw[arrow] (MADER) -- (SOGM);
    \draw[arrow] (SOGM) -- (EGOSwarm2) node[near end, above] { \cite{wu2024decentralized}};
    \draw[arrow] (EGOSwarm2) -- (SOGM);
    
    \draw[arrow] (AMSwarmX)    -- (AMSwarm)      node[midway, above] { \cite{adajania2024amswarmx}};
    \draw[arrow] (AMSwarmX)    -- (LSC)      node[midway, above] { \cite{adajania2024amswarmx}};
    
    \draw[arrow] (EDG-Team) -- (MADER) node[near start, above] { \cite{hou2022enhanced}};
    \draw[arrow] (EDG-Team)    -- (EGO-Swarm)      node[near start, above] { \cite{hou2022enhanced}};

    \draw[arrow] (EGO-Swarm) to[out=30,in=350]  node[near start, above] { \cite{zhou2021ego}} (ORCA);

    \draw[arrow] (RSFC.east) to[out=60,in=358]  node[midway, above] { \cite{zhou2021ego}} (ORCA);

    \draw[arrow] (LSC.east) to[out=45,in=150]  node[midway, above] { \cite{park2022online}} (EGO-Swarm);
    \draw[arrow] (LSC) to[out=60,in=310]  node[near end, above] { \cite{park2022online}} (BVC);
    \draw[arrow] (LSC) -- (DMPC)   node[near start, above] { \cite{park2022online}};

    \draw[arrow] (AMSwarm) -- (DPDS)   node[near start, above] { \cite{adajania2023amswarm}};
    \draw[arrow] (AMSwarm) -- (DMPC)   node[midway, above] { \cite{adajania2023amswarm}};

    \draw[arrow] (BVC) -- (ORCA)   node[midway, above] { \cite{zhou2017fast}};

    \draw[arrow] (ORCA) -- (BVC);

    \draw[arrow] (DMPC.west) to[out=125,in=15]  node[midway, above] { \cite{luis2020online}} (BVC);
    \draw[arrow] (DMPC) -- (ORCA)   node[near start, above] { \cite{zhou2021ego}};

    \draw[arrow] (DPDS) -- (BVC)   node[near start, above] {\cite{soria2021distributed}};
    \draw[arrow] (DPDS) -- (DMPC)  node[near start, above, sloped] {\cite{soria2021distributed}};

    \draw[arrow] (Ours)      -- (HDSM);
    \draw[arrow] (Ours)      -- (EGOSwarm2);
    \draw[arrow] (Ours)      -- (CAN-RL);
    \end{tikzpicture}%
  }
  \caption{
  \textbf{Comparison of multi-robot navigation algorithms.} We begin by reviewing studies that consider realistic obstacle perception and construct a comparison graph grounded in the algorithms examined across these works.
  A directed edge indicates that the source algorithm outperformed the destination algorithm in certain experiments reported in the cited paper. 
  \colorbox{forestgreen!50}{\phantom{\rule{12pt}{4pt}}} denotes control-based methods, and 
  \colorbox{darkorange!50}{\phantom{\rule{12pt}{4pt}}} denotes planning-based methods. 
  We omit SGBA \cite{mcguire_2019} from the comparison since its objective is limited to collision avoidance and exploration, without addressing goal-reaching.
  }
  \label{fig:citation_graph}
  \vspace{-2em}
\end{figure}

%% file: Sections/Background.tex
\input{Tables/notation_table}
\section{Preliminaries and Background}
\subsection{Notation}
In this paper, vectors are denoted by bold letters, such as $\boldsymbol{x}$, which describes the robot’s state, and their time‐derivatives by $\dot{\boldsymbol{x}}$. 
We consider the following quadrotor dynamics model:
\begin{align}
  \boldsymbol{a} &= \boldsymbol{g}
    + \tfrac{1}{m}\,\boldsymbol{R}\,\boldsymbol{f}_{\mathrm{thrust}}
    - \tfrac{1}{m}\,\boldsymbol{f}_{\mathrm{drag}},\\
  \dot{\boldsymbol{\omega}} &= \boldsymbol{I}^{-1}\bigl(\boldsymbol{\tau}
    - \boldsymbol{\omega} \times (\boldsymbol{I}\,\boldsymbol{\omega})\bigr),\\
  \dot{\boldsymbol{R}} &= \boldsymbol{R}\,\boldsymbol{\omega},
\end{align}
where
\begin{itemize}
  \item $m$ and $\boldsymbol{I}\in\mathbb{R}^{3\times3}$ are the mass and inertia matrices,
  \item $\boldsymbol{f}_{\mathrm{thrust}}$ and $\boldsymbol{f}_{\mathrm{drag}}$ are the thrust and drag force vectors,
  \item $\boldsymbol{\tau}\in\mathbb{R}^3$ is the total torque in the body frame, and
  \item $\boldsymbol{g}\in\mathbb{R}^3$ is the gravity vector.
\end{itemize}

\subsection{Reinforcement Learning}
We consider an on‐policy reinforcement‐learning framework based on Proximal Policy Optimization (PPO)~\cite{schulman2017proximal}. The interaction between the agent and the environment is modeled as a finite‐horizon Markov Decision Process (MDP), defined by the tuple
$
(\mathcal{X}, \mathcal{U}, f, r)
$,
where $\mathcal{X}$ and $\mathcal{U}$ denote the state and action spaces, respectively, $f(\boldsymbol{x}_{t+1}\!\mid\!\boldsymbol{x}_t,\boldsymbol{u}_t)$ represents the system dynamics, and 
$
r\colon \mathcal{X}\times\mathcal{U}\to\mathbb{R}
$
is the scalar reward function. The policy $\pi_{\theta}(\boldsymbol{u}\!\mid\!\boldsymbol{x})$, parameterized by a deep neural network with weights $\theta$, is optimized to maximize the expected discounted return over a horizon of length $T$, where $\gamma\in[0,1]$ is the discount factor:
\begin{align}
J(\theta)
&= \mathbb{E}_{\pi_{\theta}}\Biggl[\sum_{t=0}^{T}\gamma^{t}\,r\bigl(\boldsymbol{x}_t,\boldsymbol{u}_t\bigr)\Biggr], \\[1ex]
r(\boldsymbol{x},\boldsymbol{u})
&= r_{\mathrm{traj}}(\boldsymbol{x})
+ r_{\mathrm{safety}}(\boldsymbol{x}, \boldsymbol{u})
+ r_{\mathrm{efficiency}}(\boldsymbol{u})
\end{align}
The reward function balances trajectory‐tracking objectives with safety and efficiency considerations. It comprises three components
where \(r_{\mathrm{trajectory}}\) rewards accurate tracking of the reference path, 
\(r_{\mathrm{safety}}\) penalizes proximity to obstacles and other quadrotors, 
\(r_{\mathrm{efficiency}}\) penalizes excessive control effort and enforces smoothness via higher‐order derivatives of the state.

\subsection{Control Barrier Functions}
Control Barrier Functions provide a formal methodology to enforce collision constraints~\cite{ames2019control}. Given a continuously differentiable function \(h\colon \mathbb{R}^n \to \mathbb{R}\) defining the safe set
\begin{align}
\mathcal{S} = \{\,\boldsymbol{x}\in\mathbb{R}^n \mid h(\boldsymbol{x}) \ge 0\},
\end{align}
the system remains safe so long as \(h(\boldsymbol{x})\ge0\) at all times. Ensuring forward invariance of \(\mathcal{S}\) requires
\begin{align}
\dot h(\boldsymbol{x},\boldsymbol{u}) + \alpha\bigl(h(\boldsymbol{x})\bigr)\ge0,
\end{align}
where \(\alpha\) is an extended class-\(\mathcal{K}\) function.

To enforce safety for a given nominal controller \(\boldsymbol{u}_{\mathrm{nominal}}\), the following Quadratic Program (QP) is solved that minimally adjusts the policy control input:
\begin{align}
\begin{aligned}
\label{eq:safety_problem}
\min_{\boldsymbol{u}_{\mathrm{safe}}}\quad & \bigl\|\boldsymbol{u}_{\mathrm{safe}} - \boldsymbol{u}_{\mathrm{nominal}}\bigr\|^2\\
\text{s.t.}\quad
& \dot h\bigl(\boldsymbol{x},\boldsymbol{u}_{\mathrm{safe}}\bigr) + \alpha\bigl(h(\boldsymbol{x})\bigr)\ge0,\\
& \boldsymbol{u}_{\mathrm{safe}}\in\mathcal{U}.
\end{aligned}
\end{align}
This QP guarantees that the applied input \(\boldsymbol{u}_{\mathrm{safe}}\) satisfies the safety constraint defined by \(h\), while remaining as close as possible to the proposed control \(\boldsymbol{u}_{\mathrm{nominal}}\), which in our framework is generated by the learned policy.

We follow the formulation of the barrier certificate presented in \cite{wang2017safety}, where the state of the system is a concatenated vector of all robot states defined as pairwise interactions.
We consider only the 2 closest neighbors in a range of $2.0m$.
\begin{align}
    \begin{split}
         h_{i,j}^{quadrotor}(x,v) =& \sqrt{2(\alpha_i + \alpha_j)(||\Delta \boldsymbol{x}_{i,j}|| - D_s)} \\
         &+ \frac{\Delta \boldsymbol{x}_{i, j}^T}{|| \Delta \boldsymbol{x}_{i, j}||}\Delta v_{i,j}
    \end{split}
\end{align}
Where $\alpha$ denotes the maximum acceleration for each quadrotor and $D_s$ is the distance that must be enforce between a pair of quadrotors. Likewise, obstacle constraints are defined as other quadrotors (of spherical constraints) with no control authority i.e. $\alpha_j = 0$. 
\begin{align}
    \begin{split}
        h_{i,k}^{obstacle}(x,v) =& \sqrt{2(\alpha_i)(||\Delta \boldsymbol{x}_{i,k}|| - (\frac{D_s}{2} + r_k))} \\
        &+ \frac{\Delta \boldsymbol{x}_{i, k}^T}{|| \Delta \boldsymbol{x}_{i, k}||}\Delta v_{i,k}
    \end{split} 
\end{align}
Additional constraints are also added for the room boundaries, such as walls, floors and ceiling. 

%% file: Tables/notation_table.tex
\begin{table}
\centering
\caption{Notation used in this paper}
\label{tab:notation}
\resizebox{.48\textwidth}{!}{
  \begin{tabular}{|>{\centering\arraybackslash}m{.15\textwidth}|>{\arraybackslash}m{.3\textwidth}|}
    \hline
    \textbf{Notation} & \textbf{Definition} \\
    \hline
    $*_i$ & Corresponding state of the $i'th$ quadrotor \\
    \hline
    $\boldsymbol{x}, \boldsymbol{v}, \boldsymbol{a}, \boldsymbol{j}$ & Position, velocity, acceleration, and jerk. $\mathbb{R}^3$\\
    \hline
    $\boldsymbol{R}$ & Rotation matrix $\in SO(3)$\\
    \hline
    $\psi$ & quadrotor yaw \\
    \hline
    $\boldsymbol{w}, \boldsymbol{\dot{w}}$ & angular velocity, angular acceleration \\
    \hline
    $u_{policy}$ & The control action (thrusts) generated from the policy. $\mathbb{R}^4$\\
    \hline
    $u_{safe}$ & Safe control (thrusts) generated from a Barrier Function $\mathbb{R}^4$\\
    \hline
    $\mathcal{X}_{i}, \mathcal{X}_{goal,i}$ & Quadrotor state vector, evaluation of trajectory:  $[\boldsymbol{x} ~ \mathbf{v} ~ \boldsymbol{\omega}, ~ \boldsymbol{R}] \in \mathbb{R}^{18}$. \\
    \hline
    $h_{n,j}^{quadrotor}(x,v)$ & Constraint for $i'th$ quadrotor to $j'th$ quadrotor\\
    \hline
    $h_{n,k}^{obstacle}(x,v)$ & Constraint for $i'th$ quadrotor to $k'th$ obstacle\\
    \hline
    $\pi_{\theta}$ & Deep neural network $\pi$ with weights $\theta$. \\
    \hline
    $\boldsymbol{e}_i^t$ & Relative position input seen by policy: $\boldsymbol{x}_i^T - \boldsymbol{x}_{goal,i}^T$ \\
    \hline
    $\boldsymbol{\eta}_i^t$ & Relative neighbor position seen by policy: $\boldsymbol{x}_i^T - \boldsymbol{x}_j^T$ \\
    \hline
    $\boldsymbol{\zeta}_i^t$ & Obstacle ranging from multi zone time-of-flight. $\zeta \in [0.0, 2.0]m$. $\mathbb{R}^{4\times8}$ \\
    \hline
    $FC(x, y)$ & A fully connected network with inputs $x$ and output $y$. \\
    \hline
    
  \end{tabular}
  \vspace{-0.1em}
}
\end{table}

%% file: Sections/problem_statement_language.tex
\section{Problem Statement}
\label{sec:problem_statement}
We consider a 3D environment with $N$ homogeneous quadrotors and an unknown number of static obstacles with arbitrary positions and dimensions. 
Each quadrotor has no prior map of its surroundings and relies exclusively on onboard resources: power-constrained sensors for localization and perception, single‐core, sub‐gigahertz processors with less than one megabyte of memory for computation, and peer‐to‐peer communication limited to megabit‐per‐second bandwidth. Under these stringent hardware constraints, our objective is to design a decentralized navigation scheme that enables all $N$ quadrotors to reach their designated goals within a specified time limit while avoiding collisions with both obstacles and other quadrotors.

\begin{figure}[t!]
    \centering
    \includegraphics[width=0.48\textwidth]{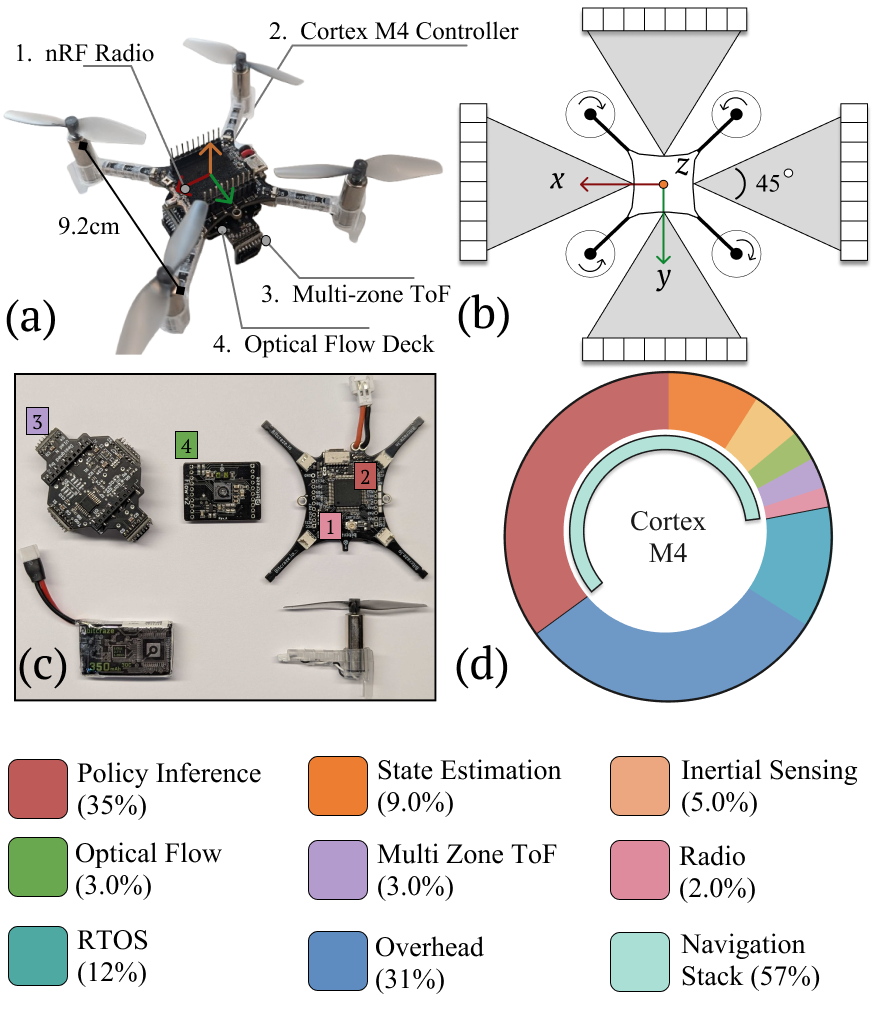}
    \caption{\textbf{Hardware System.} The Crazyflie platform used in the real world (a) and in simulation. The quadrotor is equipped with a set of 4 VL53L5CX sensors that each provide an $8\times8$ depth image (b). Each quadrotor is $9.2cm$ in size and weighs merely $47g$. We utilize the onboard nRF51822 radio to communicate neighbor positions and velocities. (c) shows the individual components where number and color denotes the corresponding compute component and (d) the breakdown of compute usage.}
    \label{fig:hardware_system}
\end{figure}

%% file: Sections/Method.tex
\section{Method}
\label{sec:method}
The navigation problem described in Section~\ref{sec:problem_statement} becomes increasingly intractable as the number of quadrotors, obstacles, and planning horizon grows. 
Solving it using only onboard, resource-constrained hardware introduces several challenges:
i) Perception and localization uncertainty due to sensor noise, limited field of view, and environmental uncertainties;
ii) Limited data exchange and communication delay arising from bandwidth constraints;
iii) Trade-off between optimality and solution times owing to limited onboard computational resources.
To overcome these challenges, this section introduces the hardware design considerations and then proceeds to the RL framework design. We believe for reliable real world results, both hardware and algorithmic considerations must be made.

\subsection{Hardware Design}
We use the Bitcraze Crazyflie 2.1, a widely adopted commercial off-the-shelf (COTS) nano-UAV platform for our experiments.
Onboard computation is performed by an STM32F405 ARM Cortex-M4 microcontroller running at 168 MHz, with 192 KB of RAM and 1 MB of flash. 
For perception, we employ the custom expansion board from Niculescu \textit{et al.}~\cite{niculescu2023nanoslam}, which carries four VL53L5CX multi-zone ToF sensors. 
Each VL53L5CX delivers either a $4\times4$ depth map at $30Hz$ or an $8\times8$ depth map at $15Hz$.
For localization, we use the Optical Flow Deck v2.
Velocity and altitude measurements are fused via an extended kalman filter to produce localization estimates for each quadrotor. Figure \ref{fig:hardware_system}(b) shows the field of view of 4 VL53L5CX sensors. 
\subsection{Algorithm Overview}
We propose a reinforcement learning (RL)-based framework that integrates trajectory planning and safety-based control to yield a fully learned, end-to-end navigation policy. The policy takes as observation the quadrotor's states, ranging data, and neighboring positions to output normalized thrusts. Unlike conventional learning-based methods that directly utilize the final goal to form observations~\cite{riviere2020glas, huang2024collision}, our approach employs a naive trajectory planner to generate smooth goal points at each timestep. These intermediate goals then serve as input for constructing the observation space. The key idea is that the trajectory generation does not account for obstacles and other quadrotors, instead, the policy is trained to modify the controls online to account for naivety. During training, we leverage privileged information for two components: a recurrent multi-head attention critic and a safety filter based on control barrier functions \cite{wang2017safety}. The safety filter is activated through a two stage training process. When training from scratch, the filter provides no useful guidance for learning how to fly. Therefore, the barrier function based reward is applied only in the latter stages of training to improve collision performance.
In the following section, we discuss scenario design (Section \ref{sec:scenario}), trajectory generation (Section \ref{sec:traj_gen}), and the reinforcement learning framework (Section \ref{sec:rl_alg}).

\subsection{Scenario design}
\label{sec:scenario}
The training environment is configured as an \(8\,\mathrm{m} \times 8\,\mathrm{m} \times 3\,\mathrm{m}\) room containing multiple static obstacles extending from the floor to the ceiling. At the start of each episode, a central obstacle region measuring either \(6\,\mathrm{m} \times 4\,\mathrm{m}\) or \(4\,\mathrm{m} \times 6\,\mathrm{m}\) is selected with equal probability. This region is discretized into 24 square grid cells, each with an area of \(1\,\mathrm{m}^2\). Cylindrical obstacles with diameters randomly chosen between \(0.2\,\mathrm{m}\) and \(0.85\,\mathrm{m}\) are placed within these cells. Each obstacle is positioned so that any point on it is at least \(0.075\,\mathrm{m}\) away from the cell boundaries, ensuring a minimum clearance of \(0.15\,\mathrm{m}\) between obstacles and resulting in at least \(0.05\,\mathrm{m}\) of traversable free space for the quadrotors.
The obstacle density, defined as the fraction of occupied grid cells, is randomly selected between 0.1 and 1.0 for each episode. Quadrotors are initialized at random positions along one \(4\,\mathrm{m}\)-side of the obstacle region, with their goal positions randomly assigned along the opposite side. Two goal assignment strategies are used: (i) all quadrotors share the same goal position, or (ii) each quadrotor is assigned a unique goal position.

During training, we choose not to terminate episodes upon collisions. We believe that the collision interaction is important for learning navigation behaviors, especially with a safety filter based guidance. Therefore, collisions between quadrotors and obstacles are explicitly simulated with randomized interactions. When two quadrotors collide, each quadrotor’s velocity direction is reversed relative to their mutual positions, and random noise is added to capture realistic collision dynamics. The speeds of both quadrotors are then reduced by $20-80\%$ of their pre-collision values. Additionally, each quadrotor’s angular velocity vector is reassigned: a random direction is sampled uniformly in 3D, and the magnitude is sampled uniformly, $\|\omega_i\| \sim  \mathcal{U}(10\pi,20\pi)$.
For quadrotor to obstacle collisions, we reverse the velocity direction with respect to the obstacle center. In this case, the new angular velocity direction is also random in 3D, with the magnitude sampled as $\|\omega_i\| \sim \mathcal{U}(\frac{\pi}{2},\pi)$ radians per second to balance the performance between passing through narrow gaps and collision avoidance.

\subsection{Trajectory Generation}
\label{sec:traj_gen}
We adopt the minimum‐snap trajectory generation method introduced by Mellinger et al.~\cite{mellinger2011minimum} for trajectory generation. At the start of each episode, we randomly sample initial and final positions for all $N$ quadrotors and compute a naive, minimum‐distance straight–line trajectory, ignoring obstacles.  Each trajectory is then cast as a piecewise polynomial under the minimum‐snap formulation.  At every timestep, we evaluate these polynomials to extract the current goal state, $\mathcal{X}_{goal} := [\boldsymbol{x},\,\boldsymbol{v} ,\, \boldsymbol{\omega},\, \boldsymbol{R}]$, which indicates position, velocity, angular velocity, and rotation matrix.

\subsection{Reinforcement Learning Framework}
\label{sec:rl_alg}
In this section, we discuss 
i) The observation and action space (Section \ref{sssection:o_a}), 
ii) reward function (Section \ref{sssection:reward}), and
iii) model architecture (Section \ref{sssection:model_arch}), 
\begin{figure*}[t!]
\centering
  \includegraphics[width=0.97\textwidth]{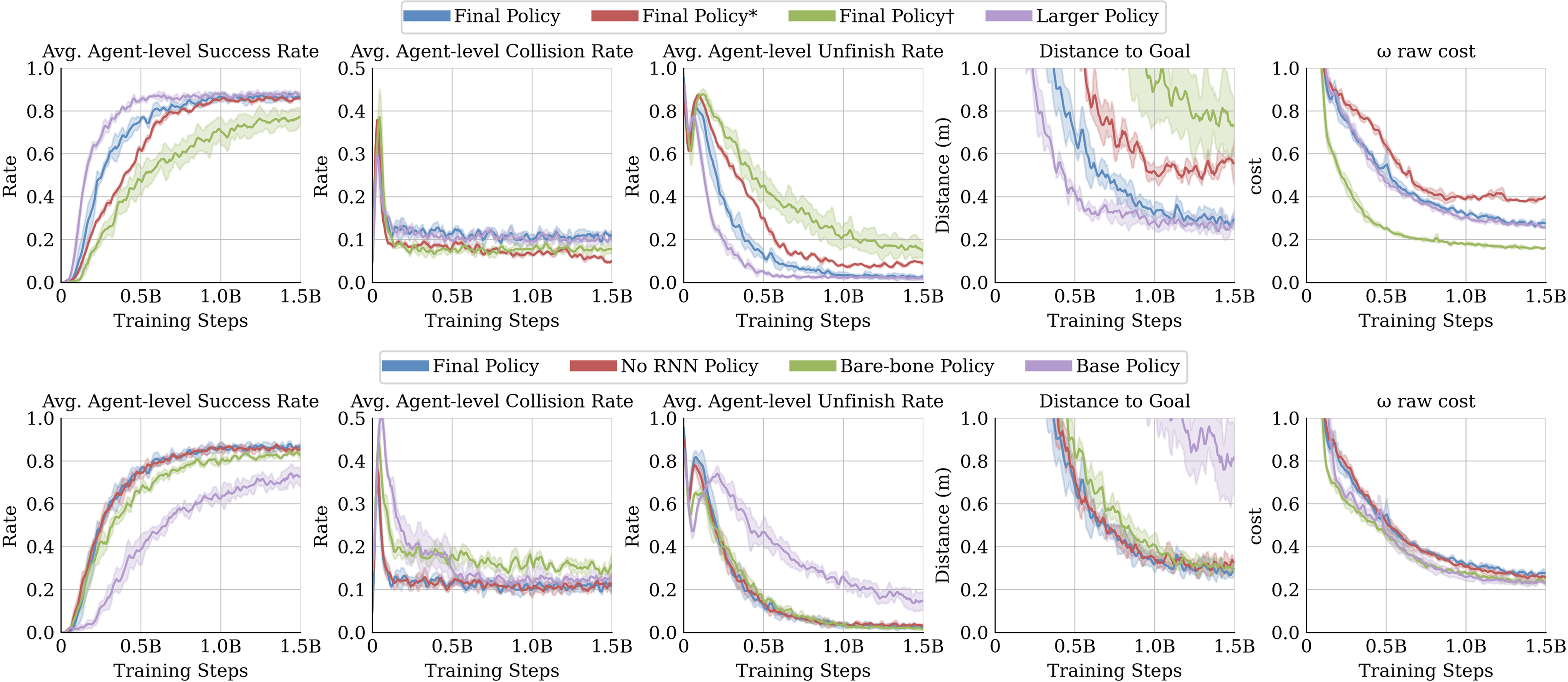}
  \caption{\textbf{Training Curves.} PPO training curves for different model variants are shown above. The final policy shown by the blue curve is the one used for all experiments, including real world and simulation. * is the final policy trained using Population Based Training \cite{jaderberg2017populationbasedtrainingneural}. † denotes a version where the safety reward is for all steps (single stage). The larger policy employs a hidden size of $64$, as opposed to $32$ in main experiments. The base policy from \cite{huang2024collision} is also compared as a baseline.}
  \label{fig:training}
\end{figure*}
\subsubsection{Observation and Action Space}
\label{sssection:o_a}
We train our policy using PPO~\cite{schulman2017proximal}, with asynchronous network architectures between the actor and critic.
At each timestep $t$, quadrotor $i$ receives observation $\boldsymbol{o}_{i, a}^t = (\boldsymbol{e}_i^t, \boldsymbol{\eta}_i^t, \boldsymbol{\zeta}_i^t),$ while the critic receives: $\boldsymbol{o}_{i, c}^t = (\boldsymbol{e}_i^t, \boldsymbol{\eta}_i^t, \boldsymbol{\xi}_i^t)$.
Here, $\boldsymbol{e}_i^t$ denotes the quadrotor’s observation of its own state and goal, $\boldsymbol{\eta}_i^t$ represents observations of neighboring quadrotors, and $\boldsymbol{\zeta}_i^t$ (actor, time of flight) and $\boldsymbol{\xi}_i^t$ (critic, signed distance field) denote observations of obstacles. 
\paragraph{Self and Goal Observation}
The quadrotor’s own state and goal observation is defined as 
$\mathbf{e}_i^t = (\mathbf{rx}_i^t, \mathbf{rv}_i^t, \mathbf{rR}_i^t, \boldsymbol{r\omega}_i^t, \mathbf{v}_i^t, \boldsymbol{\omega}_i^t)$
where 
$\mathbf{rx}_i^t \in \mathbb{R}^3$ is the position of the quadrotor relative to its goal,
$\mathbf{rv}_i^t \in \mathbb{R}^3$ is the quadrotor’s linear velocity relative to the goal’s velocity,
$\mathbf{rR}_i^t \in \mathrm{SO}(3)$ is the quadrotor’s orientation relative to the goal’s orientation,
$\boldsymbol{r\omega}_i^t \in \mathbb{R}^3$ is the angular velocity of the quadrotor relative to the goal's angular velocity,
$\mathbf{v}_i^t \in \mathbb{R}^3$ is the quadrotor’s linear velocity, and
$\boldsymbol{\omega}_i^t \in \mathbb{R}^3$ is the quadrotor’s angular velocity.

\paragraph{Neighbor Observations}
The observation of neighboring quadrotors is given by
$\eta_{i}^{t} = (\tilde{\boldsymbol{x}}_{i,1}^t, \ldots, \tilde{\boldsymbol{x}}_{i,K}^t, \, \tilde{\boldsymbol{v}}_{i,1}^t, \ldots, \tilde{\boldsymbol{v}}_{i,K}^t)$,
where $\tilde{\boldsymbol{x}}_{i,j}^t \in \mathbb{R}^3$ and $\tilde{\boldsymbol{v}}_{ij}^t \in \mathbb{R}^3$ denote, respectively, the position and velocity of the $j$-th nearest neighbor relative to quadrotor $i$. Where, $K \leq N-1$ is the number of neighbors with which the quadrotor can communicate.

\paragraph{Obstacle Observations}
For the actor, obstacle observations $\boldsymbol{\zeta}_i^t$ are constructed from hardware measurements. Each quadrotor is equipped with four ToF sensors, each producing an $8\times8$ depth map with a $45^\circ$ field of view. As all obstacles are static and extend from floor to ceiling, we condense each $8\times8$ map along the vertical axis by taking the middle row across the $z$-dimension, resulting in a $4\times8$ representation across all sensors.
For the critic, we follow the approach in~\cite{huang2024collision}: $\boldsymbol{\xi}_i^t$ is a $3\times3$ grid of minimum distances to the nearest obstacles, scaled and discretized to a pre-defined resolution ($0.1m$ in our experiments). This formulation presents a permutation invariant critic with privileged environment information, helping the policy reason against moving in directions of sensor blind spots. 

\paragraph{Action Space}
The policy outputs an action $\boldsymbol{u}_i^t \in [0, 1]^4$ for quadrotor $i$ at time $t$, specifying the normalized thrust levels for each of the four rotors. These are linearly mapped to physical thrust values, where $0$ corresponds to zero thrust and $1$ corresponds to maximum thrust.

\subsubsection{Reward Function}
\label{sssection:reward}
The total reward is formulated as three core objectives: accurately tracking a trajectory ($r_{\mathrm{trajectory}}$), maintaining a safe distance between objects and other quadrotors ($r_{\mathrm{safety}}$), and minimizing unnecessary control efforts ($r_{\mathrm{efficiency}}$): 
\[
r(\boldsymbol{x}, \boldsymbol{u})
= r_{\mathrm{trajectory}}(\boldsymbol{x}) + r_{\mathrm{safety}}(\boldsymbol{x}, \boldsymbol{u}) + r_{\mathrm{efficiency}}(\boldsymbol{u})
\]
\paragraph{Trajectory Tracking}
The trajectory tracking reward encourages the quadrotor to track a full state trajectory ($\mathbf{x} = [\boldsymbol{x}, ~ \mathbf{v},~ \boldsymbol{\omega}, ~ \mathbf{R}]$). A hybrid position penalty is applied with the following formulation: for distances greater than $0.2m$, we apply a clipped Euclidean cost; for distances within $0.2\,\mathrm{m}$, we apply a smooth exponential decay to encourage convergence. Rotation misalignment is penalized by comparing only the relative yaw. A spin penalty is also included to suppress high angular velocities such as wobbling behavior and promotes smoother flight.
\paragraph{Safety}
The safety term $r_{\mathrm{safety}}$ penalizes crashes, low-altitude flight, and proximity to constraint boundaries. To discourage flight in regimes where ground effect and localization errors become significant, a fixed crash penalty is applied when the quadrotor is detected to be on the floor, and a continuous penalty is added if the altitude falls below $0.2m$. We additionally incorporate safety-guided control (SBC) costs. These include a thrust disagreement penalty, which measures the deviation between the policy’s output and the SBC provided thrusts, scaled by the distance to constraint boundaries (i.e. $h^{\mathrm{quadrotor}_{n,j}}$ and $h^{\mathrm{obstacle}_{n,k}}$. A boundary proximity cost penalizes the quadrotor when it nears the edge of the safe set given that a solution to eq \ref{eq:safety_problem} exists. In the case the optimization is infeasible, a no-solution penalty is triggered to penalize actions that brought the quadrotors into this configuration. To ensure that safety takes precedence in high-risk states, costs relevant to efficiency are temporarily down-weighted when SBC penalties are active. This formulation allows us to balance the explorative freedom of the RL framework with the guidance from a \textit{when needed} safety filter.
\paragraph{Efficiency}
The efficiency term $r_{\mathrm{efficiency}}$ minimizes unnecessary control effort by penalizing the $L_2$ norm of the rotor thrust vector. The final reward is computed as the negative weighted sum of all cost components, scaled by the simulation timestep, $\Delta t=0.005$.
\input{Tables/ablation}
\paragraph{Model Architecture}
\label{sssection:model_arch}
The full architecture is shown in Figure \ref{fig:system_overview}. The encoders are designed to process quadrotor dynamics, neighbor, and obstacle observations in a lightweight fashion. Each observation mode is embedded via independent and asynchronous MLPs, followed by a shared single head attention mechanism and feedforward action generation. 
The actor and critic networks share the overall structure but differ in the obstacle encoder where privileged critic information, a signed distance field, is used. Furthermore, the critic employs a recurrent architecture and larger multi-head attention mechanism, as opposed to the memory-less single-head attention used for the actor.

\paragraph{Sim-to-real Deployment}
As we do not have access to a precise model that will distinctly capture $N$ different quadrotors, we apply perturbations to the dynamics, observations, and actions. For deployment, this is sufficient in generalizing to a large group, without needing re-identification across specific quadrotors. 

The model is deployed completely onboard the MCU that drives the Crazyflie. As the MCU only contains one core, the implementation is done \textit{in-series} with a Real Time Operating System (RTOS). Since we train the policy with asynchronous observation updates i.e. observations come in at a different frequencies, we forward pass encoders \textit{asynchronously} on hardware. To accomplish this, each encoder is implemented as a RTOS task with their respective frequencies, as denoted in Figure \ref{fig:system_overview}. In order to protect the information during encoder propagation, components of the model are wrapped in a semaphore. Obstacle observations ($\boldsymbol{\zeta}_i$) are captured using a buffering scheme to ensure data completeness. First, the raw unfiltered data is queried from the sensor and placed into a buffer at a rate of $15hz$. When new data is placed into the buffer, we clip ($[0m, 2.0m]$), transform ($\mathbb{R}^{4 \times8\times8} \rightarrow \mathbb{R}^{4\times8\times1}$ ) and apply an exponential filter to ready the data for the obstacle encoder. It is important to note that we do not simulate ray tracing of neighboring quadrotors in training. As a result ranging in the real world is clipped to $2.0m$ if the field of view is intersecting another quadrotor. Neighbor observations ($\boldsymbol{\eta}_i$) are exchanged using the onboard radio in an asynchronous fashion. Each quadrotor broadcasts its position and velocity at $50hz$. When another quadrotor receives this data, it compares the received position with the two nearest stored positions. The two closest received positions (and their respective velocities) are used to propagate the neighbor encoder. The entire pipeline leaves a computational overhead of $30\%$ to the microcontroller. Surrounding literature \cite{liu1973scheduling} points out that this is the upper limit to maintain the scheduling deadlines (i.e. encoder update rates) that we train the model on.

%% file: Tables/ablation.tex
\begin{table*}[!t]
\centering
\caption{\textbf{Ablation Studies.} We evaluate different variations of the policy training along with ablated components. \colorbox{cyan!25}{\phantom{\rule{12pt}{4pt}}} denotes performance gains from the ablation, where \colorbox{orange!25}{\phantom{\rule{12pt}{4pt}}} shows the trade-off.}
\label{tab:ablation}
\begin{threeparttable}
\resizebox{\textwidth}{!}{%
    \begin{tabular}{@{}llccccccc@{}}
        \toprule
        \textbf{Model Type} & \textbf{Scenario} 
        & \textbf{Quadrotor‐level $\uparrow$} 
        & \textbf{Overall $\uparrow$} 
        & \textbf{Incomplete $\downarrow$} 
        & \textbf{Q–Q coll. $\downarrow$} 
        & \textbf{Q–O coll. $\downarrow$} 
        & \textbf{Avg. Dist. $\downarrow$} 
        & \textbf{Avg. Vel. $\uparrow$} \\
        \midrule
        \textbf{LEARN}
        & Straight Line 
        & 0.975  & 0.880 & $0.0 \pm 0.0$  
        & $0.020 \pm 0.069$ & $0.020 \pm 0.058$ 
        & $12.447 \pm 0.226$ & $0.495 \pm 0.007$ \\
        
        & Swap Goal 
        & 0.843  & 0.480 & $0.240 \pm  0.431$              
        & $0.100 \pm 0.184$  & $0.083 \pm 0.112$  
        & $13.395 \pm 0.721$ & $0.526 \pm 0.024$ \\
        \midrule
        \addlinespace

        LEARN\tnote{*} 
        & Straight Line 
        & 0.622  & \cellcolor{orange!25} 0.061 & $0.939 \pm 0.242$  
        & \cellcolor{cyan!25}$0.005 \pm 0.036$ & \cellcolor{cyan!25} $0.064 \pm 0.109$ 
        & \cellcolor{orange!25} $14.921\pm 0.442$ & \cellcolor{cyan!25} $0.587 \pm 0.017$ \\
        
        & Swap Goal 
        & 0.605  & \cellcolor{orange!25} 0.020 & $0.940 \pm 0.240$              
        & \cellcolor{cyan!25}$0.026 \pm 0.077$  &  \cellcolor{cyan!25}$0.087 \pm 0.105$  
        & \cellcolor{orange!25} $16.889 \pm 0.637$ & \cellcolor{cyan!25} $0.553 \pm 0.017$ \\
        
        \addlinespace

        LEARN\tnote{†} 
        & Straight Line 
        & 0.857 & 0.367 & \cellcolor{orange!25}$0.571 \pm 0.500$  
        & \cellcolor{cyan!25}$0.0 \pm 0.0$ & \cellcolor{cyan!25}$0.025 \pm 0.051$ 
        & $12.985 \pm 0.282$ & $ 0.428 \pm 0.001$ \\
        
        & Swap Goal 
        & 0.824  & 0.306 & \cellcolor{orange!25}$0.633 \pm 0.487$              
        & \cellcolor{cyan!25}$0.010 \pm 0.050$  & \cellcolor{cyan!25}$0.043 \pm 0.070$  
        & $13.459 \pm 0.421$ & $0.444 \pm 0.015$ \\
        
        \addlinespace

        LEARN\tnote{‡} 
        & Straight Line 
        & 0.673 & 0.020 & \cellcolor{orange!25}$0.980 \pm 0.143$  
        & \cellcolor{cyan!25}$0.005 \pm 0.036$ & \cellcolor{cyan!25}$0.051 \pm 0.080$ 
        & $12.574 \pm 0.235$ & $0.494 \pm 0.008$ \\
        
        & Swap Goal 
        & 0.610  & 0.040 & \cellcolor{orange!25}$0.960 \pm 0.198$              
        & \cellcolor{cyan!25}$0.038 \pm 0.095$  & \cellcolor{cyan!25}$0.135 \pm 0.131$  
        & $13.871 \pm 0.566$ & $0.445 \pm 0.019$ \\
        
        \addlinespace
        
        No SBC Policy\tnote{1} 
        & Straight Line 
        & \cellcolor{orange!25} 0.806  & 0.327 & $0.061 \pm 0.242$  
        & \cellcolor{orange!25}$0.117 \pm 0.185$ & \cellcolor{orange!25}$0.148 \pm 0.176$ 
        & $13.032 \pm 0.730$ & $0.489 \pm 0.018$ \\
        
        & Swap Goal 
        & \cellcolor{orange!25} 0.650  & 0.14 & \cellcolor{cyan!25} $0.060 \pm  0.240$              
        & \cellcolor{orange!25}$0.260 \pm 0.206$  & \cellcolor{orange!25}$0.213 \pm 0.181$  
        & $14.879 \pm 1.124$ & $0.450 \pm 0.184$ \\
        
        \addlinespace
        
        No RNN Policy\tnote{2} 
        & Straight Line 
        & 0.962 & 0.755 & $0.041 \pm 0.200$  
        & $0.010 \pm 0.050$ & $0.026 \pm 0.057$ 
        & $12.267 \pm 0.261$ & $0.479 \pm 0.011$ \\
        
        & Swap Goal 
        & 0.853  & \cellcolor{orange!25}0.340 & \cellcolor{cyan!25}$0.04 \pm 0.198$              
        & $0.058 \pm 0.111$  & $0.118 \pm 0.111$  
        & $13.555 \pm 0.616$ & $0.434 \pm 0.019$ \\
        
        \addlinespace

        Bare-bone Policy\tnote{3} 
        & Straight Line 
        &\cellcolor{cyan!25} 0.969 & 0.755 & $0.102 \pm 0.306$  
        & \cellcolor{cyan!25}  $0.0 \pm 0.0$ & $0.018 \pm 0.044$ 
        & $12.370 \pm 0.147$ & $0.477 \pm 0.011$ \\
        
        & Swap Goal 
        & 0.785 & 0.3 & $0.18 \pm 0.388$              
        & \cellcolor{orange!25}$0.115 \pm 0.159$  & $ 0.125\pm 0.145$  
        & $14.028 \pm 0.785$ & $0.437 \pm 0.021$ \\
        
        \addlinespace

        Larger Policy\tnote{4} 
        & Straight Line 
        & 0.947 & \cellcolor{orange!25}0.755 & $0.022 \pm 0.149$  
        & $0.022 \pm 0.071$ & $0.047 \pm 0.104$ 
        & $12.226 \pm 0.190$ & $0.484 \pm 0.006$ \\
        
        & Swap Goal 
        & 0.858  &\cellcolor{orange!25} 0.340 & \cellcolor{cyan!25} $0.0 \pm 0.0$              
        & $0.035 \pm 0.087$  & $0.123 \pm 0.127$  
        & $13.430 \pm 0.629$ & $0.442 \pm 0.019$ \\
        
        \addlinespace

        Base Policy\tnote{5} 
        & Straight Line 
        & 0.798 & \cellcolor{orange!25} 0.184 & \cellcolor{orange!25}$0.633 \pm 0.487$  
        & \cellcolor{cyan!25}$0.0 \pm 0.0$ & $0.076 \pm 0.075$ 
        & $14.027 \pm 0.647$ & $0.434 \pm 0.016$ \\
        
        & Swap Goal 
        & 0.825 & \cellcolor{orange!25} 0.200 & \cellcolor{orange!25}$0.520 \pm 0.505$              
        & \cellcolor{cyan!25}$0.004\pm 0.033$  & $0.094 \pm 0.103$  
        & $14.008\pm 0.676$ & $0.434 \pm 0.022$ \\
        
        \addlinespace
        
        \bottomrule
    \end{tabular}%
}
\begin{tablenotes}[flushleft]
\footnotesize
\setlength{\columnsep}{0cm}
\setlength{\multicolsep}{0cm}
  \begin{multicols}{2}
\item[*] Denotes final policy trained using population based training \cite{jaderberg2017populationbasedtrainingneural}.
\item[†] Denotes policy trained using only single stage training.
\item[‡] Denotes a harsh collision interaction within simulation.
\item[1] Denotes without SBC, with RNN Critic.
\item[2] Denotes without RNN in critic, with SBC.
\item[3] Denotes without SBC and without RNN in the critic.
\item[4] Denotes a hidden size of 64.
\item[5] Denotes the policy used in \cite{huang2024collision}.
\end{multicols}
\end{tablenotes}
\end{threeparttable}
\vspace{-1.0em}
\end{table*}

%% file: Sections/Results.tex
\input{Tables/main_comparison}
\section{Results}
We validate our system extensively in both simulation (Section \ref{sec:sim_experiments}) and in real world experiments (Section \ref{sec:hardware_experiments}), to show the performance in four main attributes: 
i.) navigation through dense obstacle environments in simulation and the real world ii.) how learning-based methods perform against traditional motion planning iii.) the robustness of the overall system and iv.) experimental insights in key points of the proposed framework. We present two definitions of success rate: quadrotor-level and overall. We define quadrotor level as the fraction of quadrotors who arrived within distance $\delta$ of the goal collision free. The overall success rate is defined as the fraction of entire runs where all quadrotors arrived within distance $\delta$ of the goal collision free. In both cases, we use a threshold distance of $\delta = 0.1m$. The incomplete rate is found by taking the fraction of runs where \textit{all} quadrotors arrive within $\delta$ of the goal and were collision free. The remaining statistics (quadrotor collisions, distance, velocity, acceleration, jerk) were all found as a fraction over the number of quadrotors.

We perform evaluations in two navigation scenarios: \textit{Straight Line} and \textit{Swap Goal} each involving 8 quadrotors and $N=50$ randomized trials. Obstacles are modeled as cylinders with radius uniformly sampled in the $[0.35,0.85] m$ range and a minimum inter‐obstacle gap of $0.25m$. At each trial, quadrotors must either traverse a long corridor (Straight Line) or swap goals across the y-axis (Swap Goal), while avoiding both obstacles and inter‐quadrotor collisions. No global map is provided and each quadrotor observes only a fixed radius around it.
\subsection{Ablation Studies}
We compare eight variants of LEARN to isolate the contributions of the safety-guided barrier framework, the privileged RNN critic, and network capacity. We also investigate the choice of single stage training and a harsher collision simulation (a wider sampling bound for collision dynamics). Table \ref{tab:ablation} shows that our full model combining a two stage safety reward scheme, recurrent multi-headed critic, and a compact attention network yields the best or near-best quadrotor-level and overall success rates, with minimal collisions.
\begin{figure*}[th!]
\begin{center}
  \includegraphics[width=0.98\textwidth]{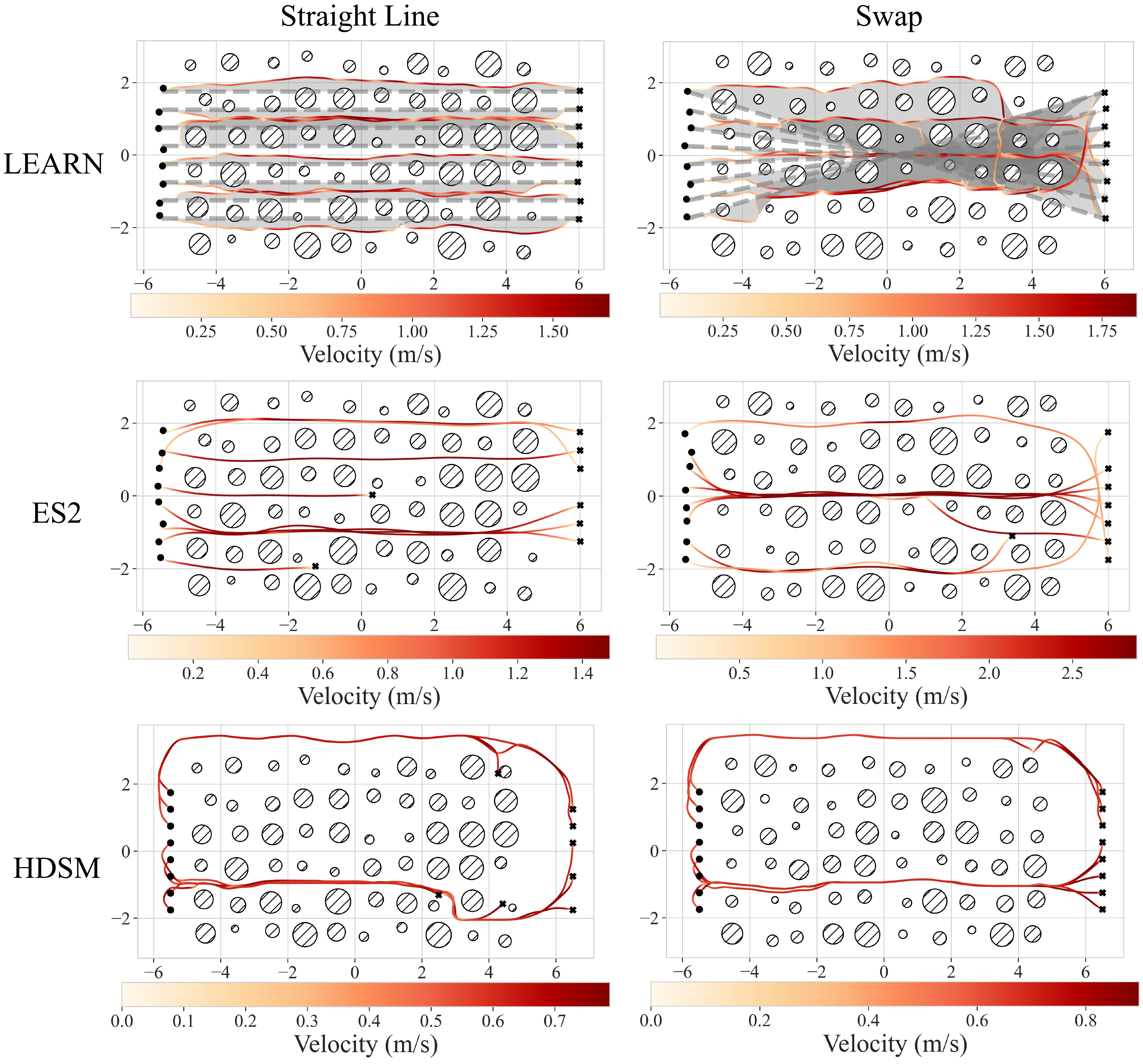}
  \caption{\textbf{Trajectory Evaluation.} Comparison between two state of the art multi quadrotor motion planning frameworks ES2 \cite{zhou2022swarm} and HDSM \cite{toumieh2024hdsm}, and LEARN in a long corridor type navigation task. The naive planned trajectories are shown as dotted lines. The velocity-colored trajectories are provided for a qualitative comparison. While LEARN has no objective function, the defined reward function often shapes how the resulting navigation path behaves. Our experiments show that LEARN successfully navigates more obstacle configurations with comparably less collisions than the two baseline comparisons.}
  \label{fig:motion_planning_comparison}
  \vspace{-1em}
\end{center}
\end{figure*}
Removing SBC causes a dramatic rise in obstacle collisions and a drop in overall success (from 0.48 to 0.14 in the Swap Goal task), underscoring its role in guiding safe exploration. Omitting the RNN critic has a milder effect (overall success falls to 0.34), indicating that memory aids navigation but is secondary to safety guidance. The minimal bare-bones policy (no SBC, no RNN) still succeeds in straight-line tasks but largely fails under goal-swap complexity. Simply increasing model size (with a hidden size of $H=64$) degrades performance, confirming that training strategies becomes more important than number of tunable parameters. Finally, the Base Policy, representing a version without architectural refinements \cite{huang2024collision}, under-performs across the board.
\subsection{Simulation Experiments}
\label{sec:sim_experiments}
Our approach is compared against Ego-Swarm 2 (ES2) \cite{zhou2022swarm} and HDSM \cite{toumieh2024hdsm} within a simulation environment. We choose the baselines given the comparisons described by Figure \ref{fig:citation_graph}. It is important to note that these two baselines present results in the real world, but are not deployable on a resource-constrained platform. Further, they rely on high dimensional sensing such as depth cameras \cite{zhou2022swarm} or LiDARs\cite{toumieh2024hdsm} to generate voxel based maps and trajectory sharing schemes. Our approach on the other hand, does not rely on map generation and uses only a lightweight communication framework.
\subsubsection{Comparison against Motion Planning}
\begin{figure*}[th!]
\centering
  \includegraphics[width=0.98\textwidth]{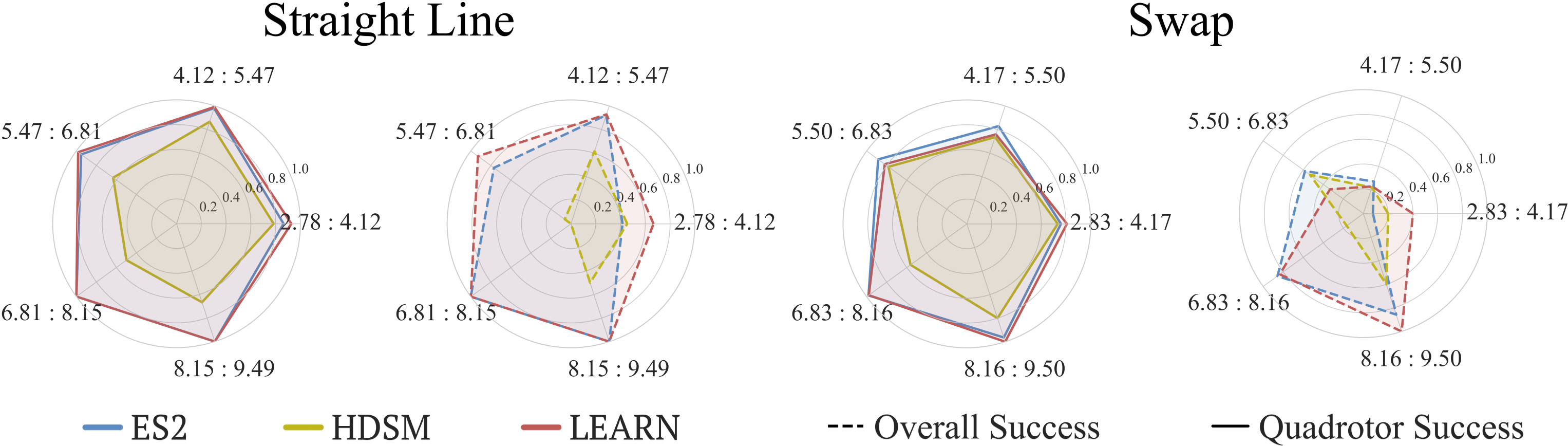}
  \caption{\textbf{Traversability.} Performance breakdowns across the multi-quadrotor traversability metric (eq. 11). The metric measures the environment difficulty against the quadrotor size ($r=0.046$ for all cases) and number of quadrotors. Each radar plot shows the range of the traversability metric as an edge. The solid lines indicate quadrotor-level success rates, whereas the dotted show overall success. A higher traversability indicates easier environments; lower traversability equates to more difficult.}
  \label{fig:traversability}
  \vspace{-1em}
\end{figure*}
\paragraph{Quantitative Results}
\begin{figure*}[t!]
\begin{center}
  \includegraphics[width=0.98\textwidth]{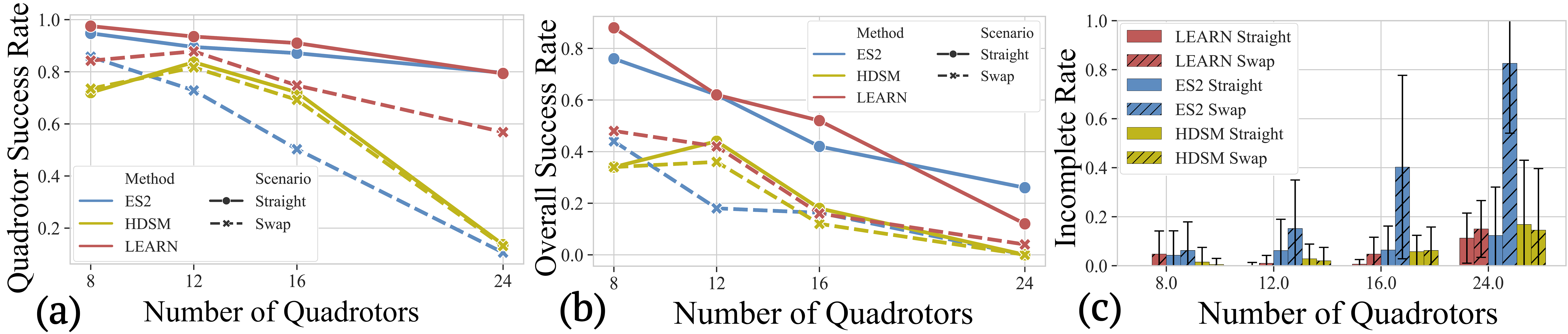}
  \caption{\textbf{Scaling Experiments.} We conduct experiments that stress the scalability against baselines. Our results present both the overall and quadrotor level success rates, along with the fraction of quadrotors who did not reach the goal (unfinished rate). We find that the motion planning baselines are unable to complete the given environments for larger number of quadrotors as compared to LEARN.}
  \label{fig:scaling_statistics}
  \vspace{-1em}
\end{center}
\end{figure*}
Table \ref{tab:comparison} provides a detailed breakdown of metrics. In the \textit{Straight Line} scenario, LEARN achieves both the highest overall success rate ($0.88$) and the highest quadrotor-level success rate ($0.975$) while completing the task fully. While ES2 performs similarly in terms of the quadrotor-level success rate, the planner fails to complete $18\%$ of trials i.e. $18\%$ contained at least one quadrotor not reaching the goal or colliding. In comparison to HDSM, our method reported a higher quadrotor to obstacle collision rate, but was able to trade-off by showcasing a much lower quadrotor to quadrotor collision rate. The overall distance traveled for our method is also the lowest against the two planners, supporting the previous qualitative claim on using the minimum distance planner. Our method also demonstrates the smoothest flight, given the lowest jerk. 

The \textit{Swap Goal} scenario was conducted to simultaneously stress the obstacle and quadrotor navigation abilities. ES2 achieves the highest quadrotor-level success rate ($0.860$), likely due to their trajectory sharing schemes allowing for denser navigation. Despite the amount of information exchanged between quadrotors in ES2, LEARN only falls behind by $0.015$. In terms of overall success rates, the policy outperforms ES2 by $5\%$, suggesting that while our method relies on much more limited information it still yields effective multi quadrotor navigation. Most importantly, we do not rely on building a voxelized representation of the environment, such as in HDSM. From these complexities, the HDSM planner suffers the most in the \textit{Swap Goal} scenario. The results show high quadrotor to quadrotor collision rates ($0.27$) and the lowest overall success rate $0.34$. We believe this is due to the way the synchronous planning is implemented, which especially suffers in tighter corridor navigation. When examining flight speeds, it is important to note that LEARN travels significantly slower than ES2. This occurs because our training environment randomly plans trajectories using a desired velocity sampled from a normal distribution, $\mathcal{N} \sim( \mu=0.5, \, \sigma=0.1$). Thus, to remain well within the distribution, the planned trajectories used in benchmarks were fixed at $v_{desired}=0.5$. We chose a normal distribution centered at $v=0.5$ to accommodate for the limits of the hardware platform. To fly faster, one can train the policy on higher velocity trajectories as done in \cite{kaufmann2023champion} to achieve higher flight speeds with a suitable platform.

Figure \ref{fig:traversability} plots success rates against traversability bins, where traversability is defined as the average unobstructed ray distance over $N$ samples normalized against quadrotor radius ($r=0.046m$ for all cases) \cite{nous2016evaluation}. In the multi quadrotor case, we propose a modified multi quadrotor traversability metric:
\begin{equation}
    \mathcal{T} = \frac{1}{j \cdot N \cdot r}\sum^N_is_i
\end{equation}
where $j$ denotes the number of quadrotors and $s_i$ the maximum ray distance (assuming full field of view). 
As traversability decreases (i.e. smaller sampled ray distances), all methods exhibit lower success rates and subsequently higher collision rates. However, our method degrades more efficiently: it sustains high quadrotor‐level and overall success even at low traversability values. In contrast, ES2 and HDSM see much steeper drops in success and sharp rises in collisions as the environment becomes more cluttered. However, we note that ES2 maintains marginally higher performance in swap goal cases due to their denser sharing scheme. 
\paragraph{Qualitative Results}
Figure \ref{fig:motion_planning_comparison} shows the \textit{Straight Line} scenario, where all quadrotors must traverse from $-x$ to $+x$ through an obstacle-dense field. Qualitatively, we find that the trajectories realized in all methods studied are visually smooth. LEARN shows a more direct traversal strategy that respects quadrotor to quadrotor distances. We believe that the minimum distance naive trajectory (the dotted line in Figure \ref{fig:motion_planning_comparison}) is the main contributor to the efficient traversal behavior. In comparison, ES2 exhibits more conservative behaviors where quadrotors tend to slow down and maintain a larger distance around obstacles. HDSM demonstrates the most conservative maneuvers that loop around to completely avoid the obstacle field.

Figure \ref{fig:motion_planning_comparison} highlight the \textit{Swap Goal} scenario, where quadrotors must traverse from $-x$ to $x$ after mirroring goal positions across the y-axis. LEARN showcases more complex behaviors, where quadrotors find respective openings in the obstacle field to avoid quadrotor to quadrotor collisions. ES2 and HDSM both showcase similar behaviors by simplifying into a single-line formation. This is likely due to the constraints imposed by the planning process that forces quadrotors to plan into the largest gap. This may become cumbersome when dealing with larger quadrotor numbers and also can result in a higher likelihood of quadrotor to quadrotor collisions. Given these trajectories, we find that our method is able to distribute the quadrotors in a sparse and safe fashion, while preventing obstacle collisions. 

\subsubsection{Scaling}
To evaluate the scalability and robustness of our navigation policy, we conducted experiments across group sizes of 8, 12, 16, and 24 quadrotors in both the \textit{Straight Line} and \textit{Swap Goal} scenarios (Figure ~\ref{fig:scaling_statistics}). The quadrotor level success rates are reported in Figure \ref{fig:scaling_statistics}(a), and the overall success rates in Figure \ref{fig:scaling_statistics}(b). In the \textit{Straight Line} task, our method exhibits strong quadrotor-level success rates, declining only from $0.975$ (8 quadrotors) to $0.882$ (24 quadrotors). The overall success rate, shows a sharper drop from $0.88$ to $0.16$. Despite this, the incomplete rate remains low and well-controlled, increasing from $0\%$ at 8 quadrotors to just $0.103$ at 24 quadrotors. This indicates that the majority of failures are due to collisions rather than inability to reach the goal. Furthermore, in the presence of a collision the policy is able to regain stability in order to reach the desired goal positions (given by the low incomplete rates). In contrast, ES2 and HDSM exhibit significant degradation at earlier quadrotor counts; for instance, HDSM's incomplete rate reaches $0.3$ at 8 quadrotors and doubles at 24 quadrotors. ES2 follows a similar relationship, doubling from a $0.2$ incomplete rate when 8 quadrotors are deployed to $0.465$ at 24 quadrotors.
\begin{figure}[t!]
\centering
  \includegraphics[width=0.48\textwidth]{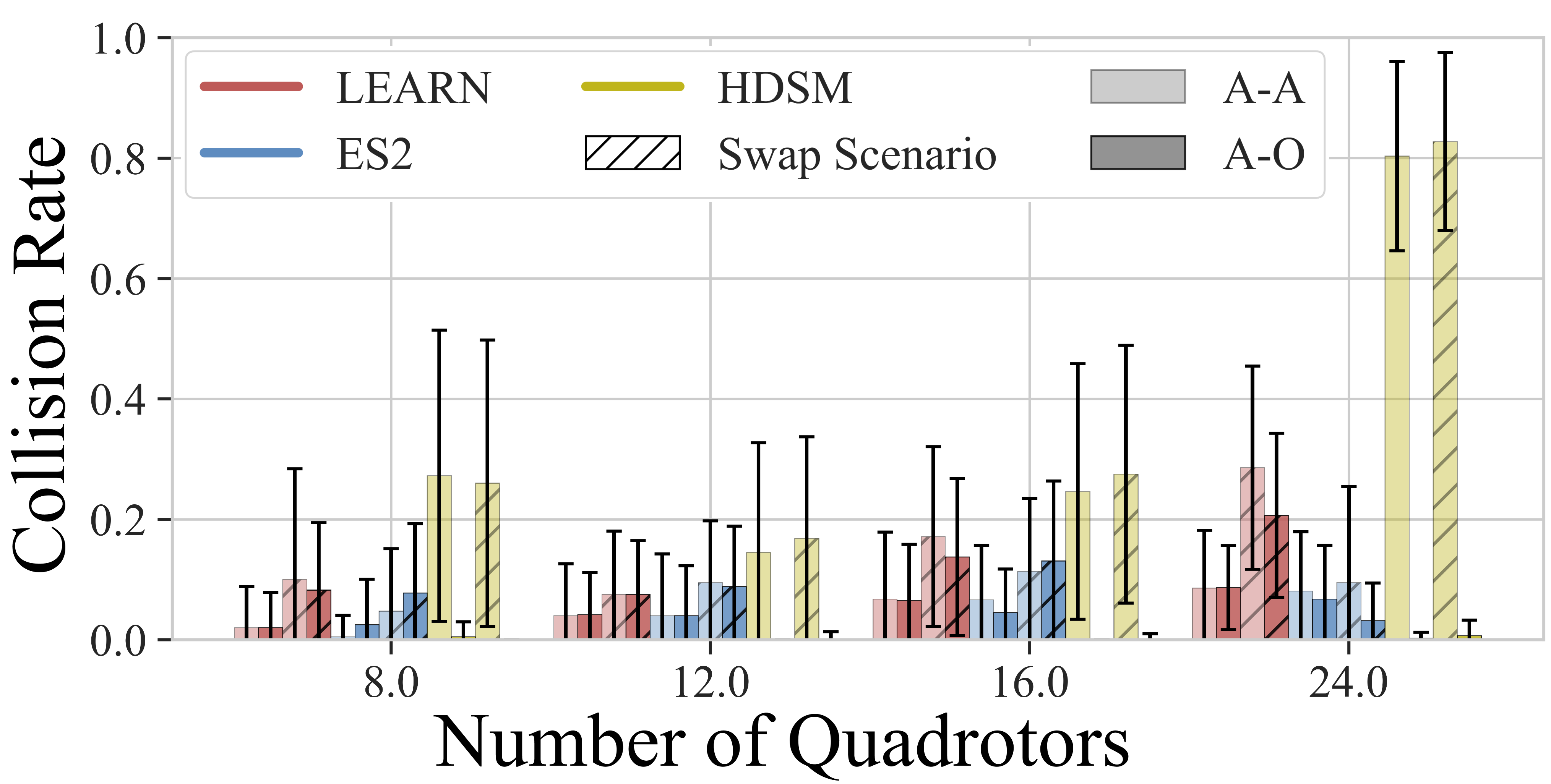}
  \caption{\textbf{Collision Scaling.} Breakdown of collision rates as the number of quadrotors scale upwards. Our method, LEARN, becomes subject to higher collision rates in comparison to ES2 when 16 quadrotors are deployed.}
  \label{fig:scaling_collision}
\end{figure}
When examining the \textit{Swap Goal} scenario, our method's quadrotor-level success begins at $0.865$ (8 quadrotors) and decreases to $0.635$ (24 quadrotors), while the overall success rate declines more steeply from $0.480$ to $0.06$. The incomplete rate correspondingly increases from $0.2$ to $0.5$, highlighting the complexity of path-crossing. Importantly, even under these difficulties, our method maintains bounded collision rates (Figure \ref{fig:scaling_collision}), with quadrotor-quadrotor collisions increasing sub-linearly $(0.089$ to $0.281$), demonstrating the policy's ability to maintain robustness. In comparison, HDSM and ES2 become bottlenecked by constraints (i.e. inter-quadrotor distance or minimum obstacle distances). The ES2 framework becomes stalled is unable to begin solving for the trajectory at 24 quadrotors, leading to the much lower collision rates. On the other hand, HDSM exhibits larger degradation in the quadrotor to quadrotor collision rates. We believe that the hyperplane separation used in the planner becomes too naive when deploying larger numbers of quadrotors. This is especially emphasized when the task involves swapping in obstacle and quadrotor rich environments.

A key differentiator lies in the communication and planning frameworks: both ES2 and HDSM rely on explicit trajectory sharing, where quadrotors broadcast their planned paths to start a joint-planning process. This approach introduces substantial bandwidth requirements, quadrotor to quadrotor temporal coupling, and excessive amounts of information, making the systems brittle in bandwidth-limited or high traffic situations. Our method, requires only exchanging instantaneous state vectors ($[x , \,\dot{x}]$) yielding a communication scheme that is both computationally and bandwidth efficient. Despite this minimalism, our method maintains higher robustness to increasing quadrotors counts, particularly in decentralized settings where full trajectory negotiation is difficult to deploy. Another reason comes from how the policy is trained. During training, the reward is a sum of \textit{all} quadrotors' performance; whereas ES2 and HDSM solve their path planning \textit{independently}. This causes an RL based approach to have higher overall success rates i.e. the policy is trained to maximize the rewards of all quadrotors not just individual. Despite this, the policy is deployed in an asynchronous fashion. 

These results collectively illustrate that our attention-based policy exhibits favorable scalability properties. The model trades off between completeness and collision rates as group density increases, outperforming classical planners in scenarios where computation, sensing, and dense communication become bottlenecks. 
\begin{figure*}[ht!]
 \centering
  \includegraphics[width=0.99\textwidth]{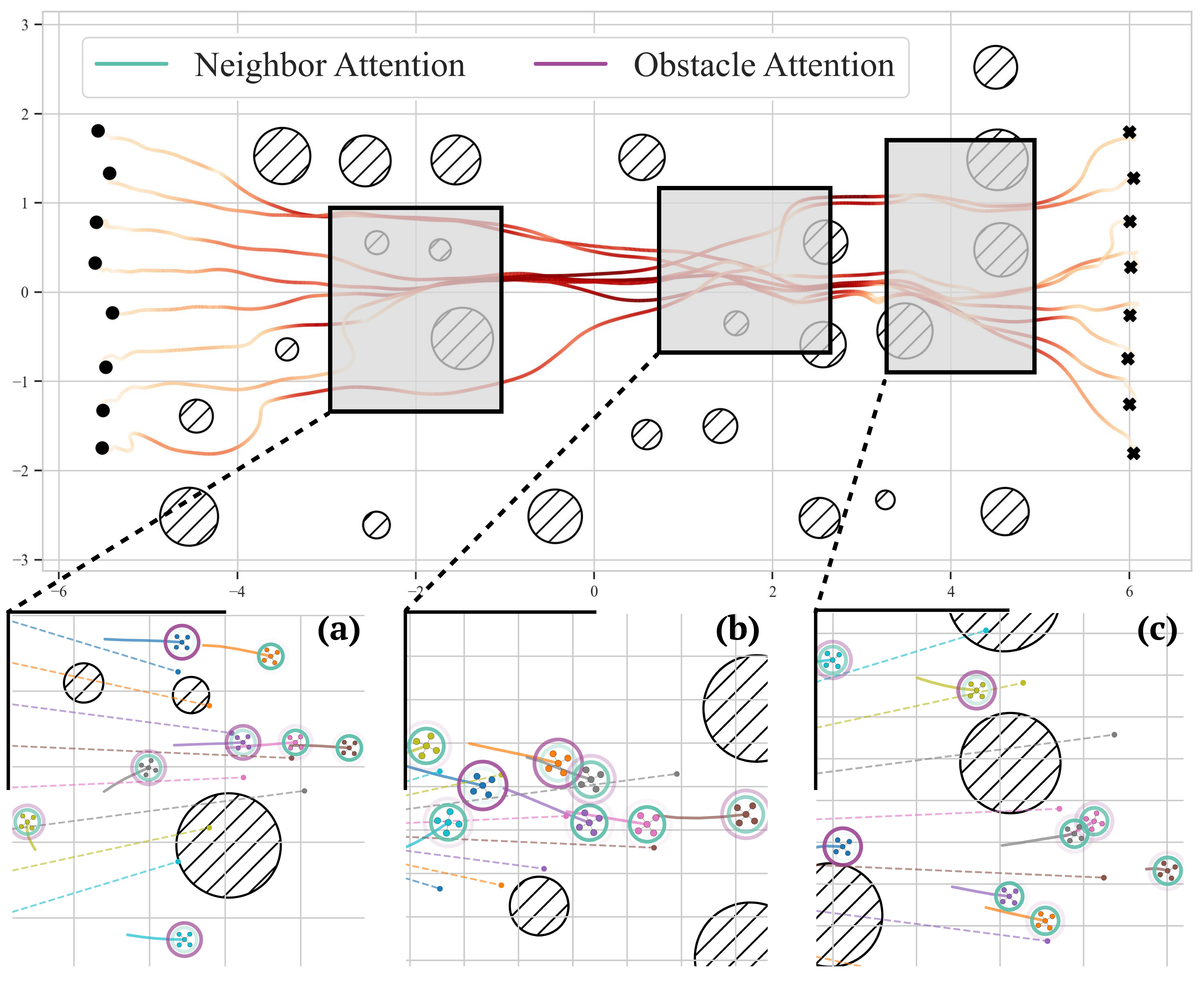}
  \caption{\textbf{Attention Analysis.} We analyze the attention mechanism on a swap goal environment. The softmax outputs for the single head attention are shown as rings around each quadrotor. The softmax represents how much emphasis the policy is placing on each of the encodings, such as balancing neighbor and obstacle observations. A bolder color represents more emphasis, while a softer color shows less importance placed. The planned naive trajectory is shown for the current time step, and a trail of the past states is plotted for reference.}
  \label{fig:attention_analysis}
\end{figure*}
\subsubsection{Robustness to Communication Delays}
We investigate the policy's robustness to reduced communication frequencies. Table \ref{table:comm} shows the two success rates, quadrotor to quadrotor collisions (Q-Q coll.), and quadrotor to obstacle collisions (Q-O coll.). We conduct over 50 trials where the state exchange rate of the environment is reduced. We observe that the performance of the policy in the straight line scenario remains above $95\%$ for all trials. Whereas, quadrotor to quadrotor and obstacle collision rates remain below $4\%$. In the \textit{Swap Goal} experiments, the performance is similarly consistent where the success rates remain above $85\%$. These results confirm that our policy tolerates significantly lower update rates, making it especially well‐suited for nano‐sized quadrotors with limited communication bandwidth. Given that our quadrotors travel at under 1.0 m/s on average, even infrequent state exchanges suffice to maintain collision free trajectories; however, abrupt velocity changes (for example, following an obstacle impact) could, at very low frequencies, lead to sudden increases in collision risk. We also believe that sharing longer horizon trajectories (common to motion planning methods), become much less reliable in low communication settings. As message exchanges become infrequent, the shared trajectories become stale much faster. This can often lead to undesirable relationship between communication and performance. 
\subsubsection{Attention Analysis}
To motivate the usage of an attention mechanism for multi quadrotor navigation, we explore the softmax outputs generated from the single head attention model (purple block in Figure \ref{fig:system_overview}). Figure \ref{fig:attention_analysis} shows the progression as the policy tracks the trajectory, avoids obstacles, and negotiates neighboring collisions for a swap goal scenario. Subfigures (a-c) represent snapshots where the attention softmax is shown as rings around each quadrotor. The softmax value represents the amount of \textit{attention} that is being placed for the corresponding inputs (neighbor or obstacle). We choose to exclude the dynamics encoding by re-normalizing against the neighbor and obstacle inputs. This is due to the observation that the dynamics attention is always prevalent and represented in attention. As we train from scratch i.e. the policy must first learn to first fly and track trajectories, this forces the policy to always attend to the dynamics. 
\input{Tables/comm_freq}
Figure \ref{fig:attention_analysis}(a) portrays the group as it navigates the beginning of the obstacle field. The yellow quadrotor in the back must first yaw to ensure that there is free space before steering around the obstacle. The balance between the green and red rings shows that the quadrotor is still mindful of others that are just making it past the obstacle, whereas the quadrotors in the lead must negotiate their positions in the obstacle free space. In Figure \ref{fig:attention_analysis}(b) we see that the group is able to balance well between both of the encoded inputs as the naive trajectories overlap. In this case, the relative positions between the quadrotor and the current trajectory evaluation ($x_{goal}$) plays an important role. When the evaluated goal is in an area of high neighbor traffic (seen in the cyan, yellow, and pink quadrotors), the policy must focus on lagging or overshooting the desired goal to prevent collisions. On the contrary, if the evaluated goal point is in the direction of obstacles (orange, blue) the policy must focus on finding an obstacle free path. Furthermore, we find exciting the ranging sensors (i.e. $\exists \, \boldsymbol{\zeta}_i<2.0 \, \forall \,i$) does not simply equate to immediately attending to the obstacle encoder. This is exhibited in the purple quadrotor of snapshot Figure \ref{fig:attention_analysis}(b). The trajectory velocity is pointing parallel to the obstacle i.e. continuing the straight line, but the quadrotor's current proximity to other neighbors forces it to focus on keeping a safe distance within the group. Finally, in Figure \ref{fig:attention_analysis}(c) we observe that the lower right group clears the obstacles and must negotiate the remaining free space among themselves. The three remaining quadrotors must still focus on the obstacles in front, yet are conscious of the others in front, such as the cyan quadrotor. 

\subsection{Hardware Experiments}
\label{sec:hardware_experiments}
\begin{figure}[t]
\begin{center}
  \includegraphics[width=0.48\textwidth]{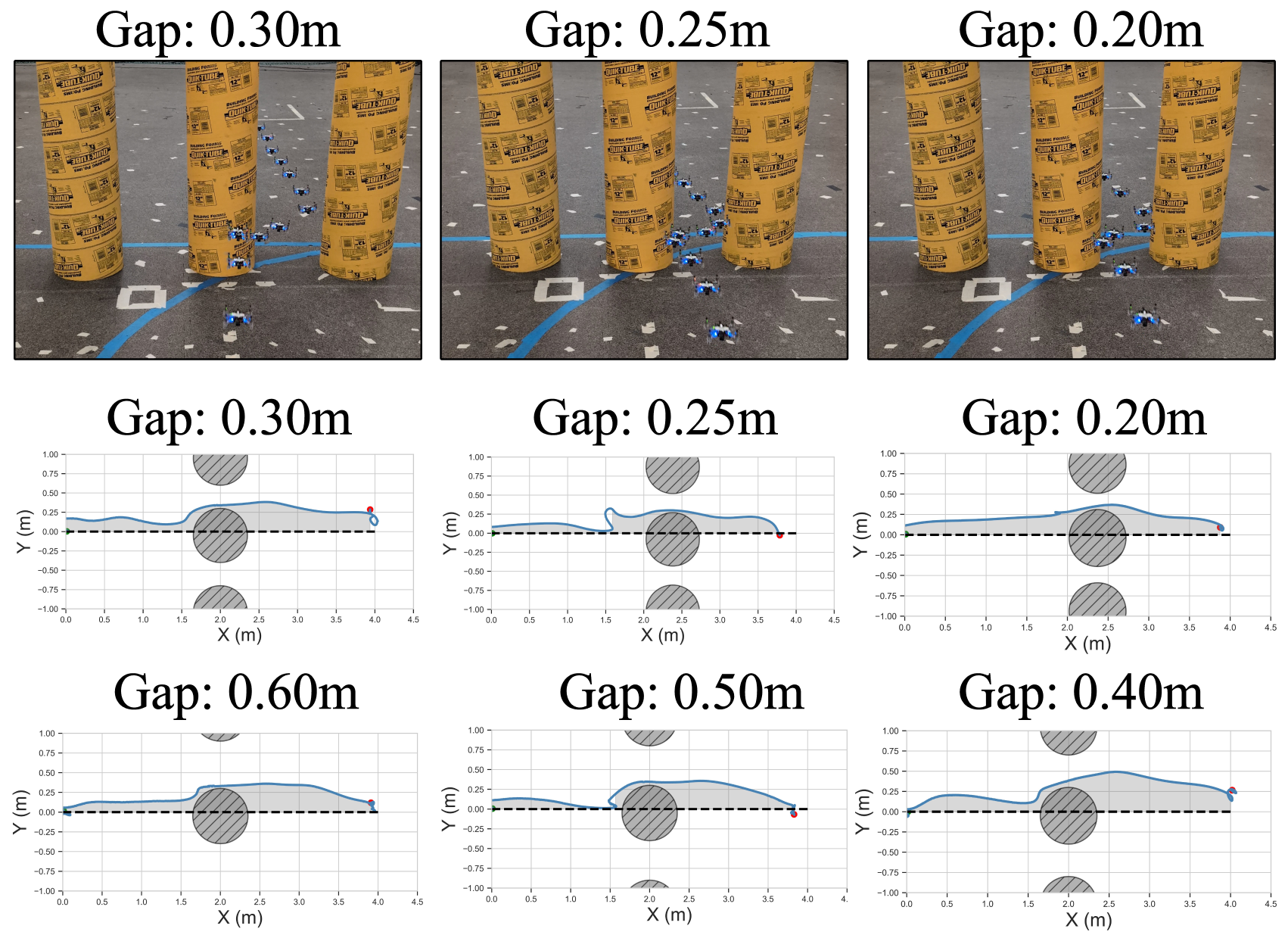}
  \caption{\textbf{Gap Test.} LEARN is tasked with flying through decreasing gap sizes. During training, we limit the environment generation to only produce gap sizes as small as $0.15m$. The real world deployment shows reliable flight down to gaps as small as $0.2m$. This experiment uses the same policy, but with ablated neighbor encoders.}
  \label{fig:gap_test}
  \vspace{-0.5em}
\end{center}
\end{figure}
We validate our policy’s real-world performance using a fleet of six Crazyflie 2.1 nano quadrotors equipped with VL53L5CX time-of-flight sensors \cite{muller2023robust}. All processing, including sensor fusion, state estimation, trajectory planning, policy inference, and communication, is performed onboard, demonstrating feasibility on highly resource-constrained platforms. Experiments are conducted in both indoor warehouse environments and outdoor settings, employing cylindrical obstacles to test the policy’s real-world generalization. Goal positions are provided through the Crazyswarm ROS environment~\cite{preiss_2017}, while execution, sensing, and collision avoidance rely entirely on each quadrotor’s onboard compute.
\subsubsection{Obstacle Avoidance}
We evaluate our policy in the real world by first tasking a single quadrotor to navigate through progressively narrower gap sizes. Below we show the trajectories of a single quadrotor flying through gaps ranging from sizes of $0.6m$ to $0.2m$, whereas the planned trajectories are shown in the dotted. The policy consistently maintains a close yet safe distance to the seen obstacles, allowing it to tightly navigate corridors. This demonstrates that the trajectory tracking reward and obstacle collision penalty work together well for this scenario. As such, our experiments indicate that we are able to reliably fly through a $0.2m$ gap, roughly 2 times the quadrotor diameter. 

\subsubsection{Multi Quadrotor Navigation}
\begin{figure*}[t]
\centering
  \includegraphics[width=0.95\textwidth]{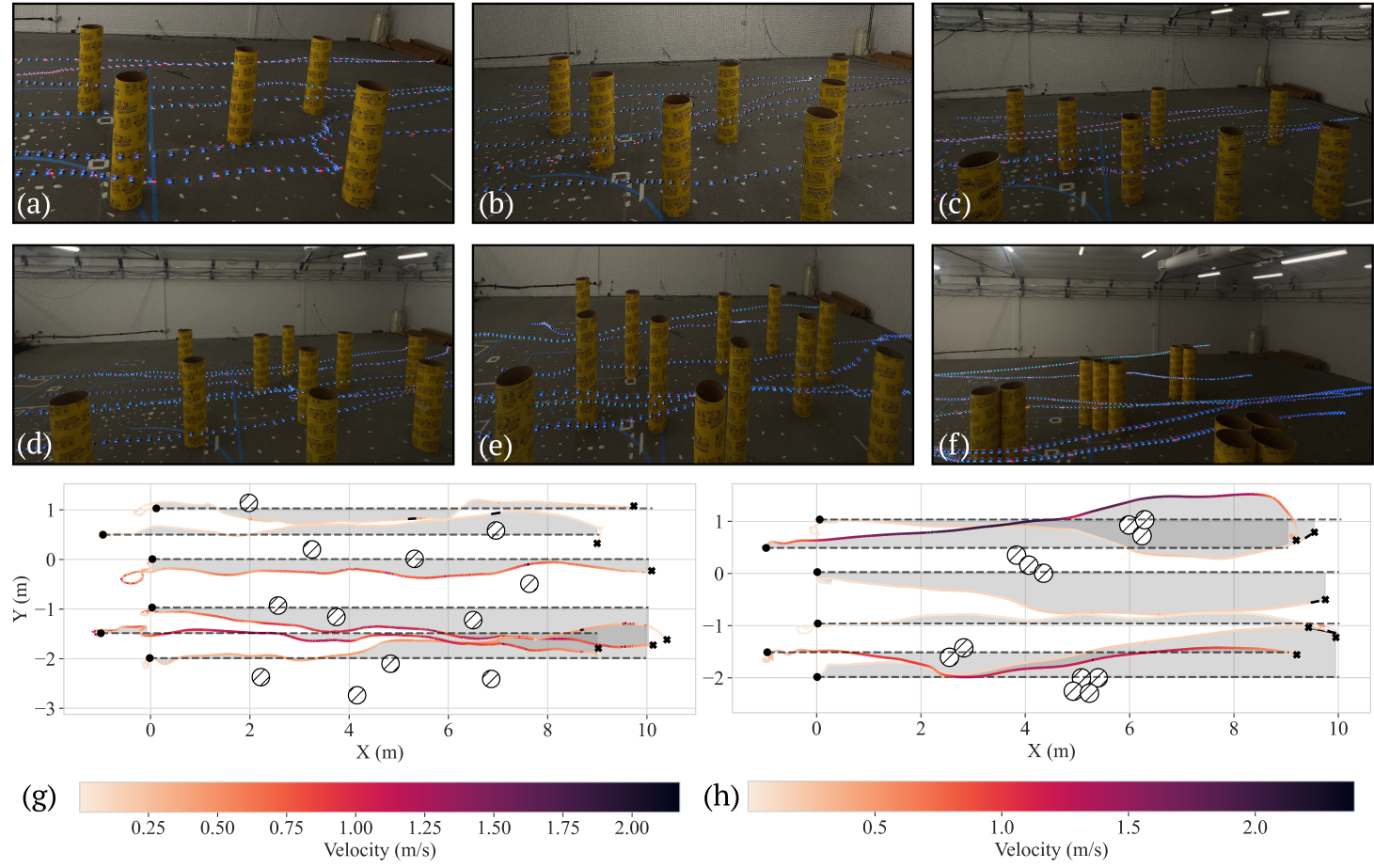}
  \caption{\textbf{Deployment Test.} Various real world deployment configurations from easy (a) to difficult (e). All individual obstacles are $0.61m$ in diameter. (f) shows that the model is able to transfer to randomized obstacle sizes. The group is tasked with tracking a $10m$ long straight line trajectory that intersects a field of obstacles. (g) represents the recorded trajectory for deployment (e). (h) is the recorded trajectory for deployment (f).}
  \label{fig:hardware_composite}
\end{figure*}

Finally, we conduct experiments by evaluating trajectory tracking, obstacle avoidance, and neighbor avoidance together. We evaluate our models in 6 randomly generated obstacle configurations in an indoor warehouse and various outdoor situations. A composite image of the systems deployed trajectories are shown in Figure \ref{fig:hardware_composite}. The difficulty of the tested environment increases with (a) being the easiest and (e) being the hardest. We group obstacles together in Figure \ref{fig:hardware_composite}(f) to show that the deployed model does not overfit to a cylindrical obstacle size. In this case the obstacle sizes range from 0.35m to 1.0m with linear diagonals. Our framework exhibits near zero collision rates with respect to both obstacles and other quadrotors. The recorded trajectories for (e) and (f) are shown in Figure \ref{fig:hardware_composite}. The policy remains close to the planned trajectory until an obstacle is seen within the field of view. \looseness=-1
\paragraph{Goal Swapping}
\begin{figure*}[th!]
\begin{center}
\includegraphics[width=0.98\textwidth]{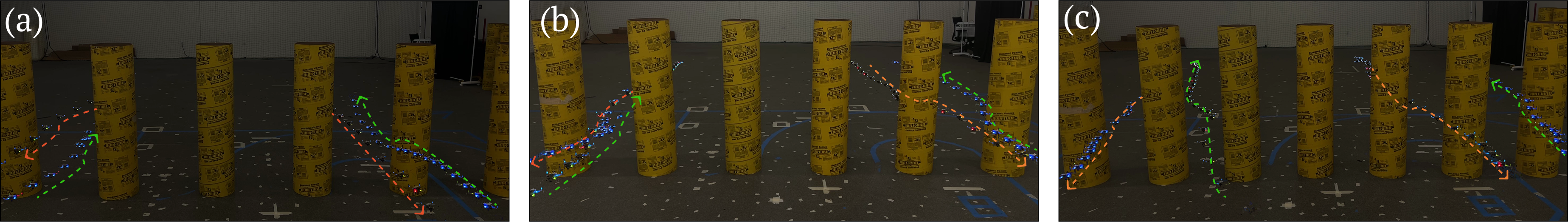}
  \caption{\textbf{Unilateral Swap.} A unilateral swap configuration. Quadrotors begin in a square and must swap positions with the quadrotor directly opposite of them. As the gap size of the obstacles grow smaller, the policy chooses to separate diverge quadrotor trajectories. (a) shows a gap size of $0.5m$, where (b) and (c) show gap sizes of $0.4m$ and $0.3m$ respectively.}
  \label{fig:unilateral_swap}
  \vspace{-0.5em}
\end{center}
\end{figure*}
We also evaluate the policy's ability in a symmetric goal-swapping task. Four quadrotors are initialized in a square, each required to reach the location directly opposite their starting position. In wider gap scenarios (e.g. $0.5m$), the quadrotors naturally select the most direct shared gap, minimizing deviation from their planned straight-line paths, even if that means multiple quadrotors simultaneously pass through the same corridor. This indicates that the policy can implicitly reason about observed obstacle locations to share space without collisions.

As the gap narrows to $0.3m$, the behavior shifts: quadrotors actively select different paths by deviating farther from their planned trajectories. This emergent behavior showcases the policy’s capacity to adapt its aggressiveness in response to environmental conditions.
\begin{figure*}[t!]
\centering
\includegraphics[width=0.98\textwidth]{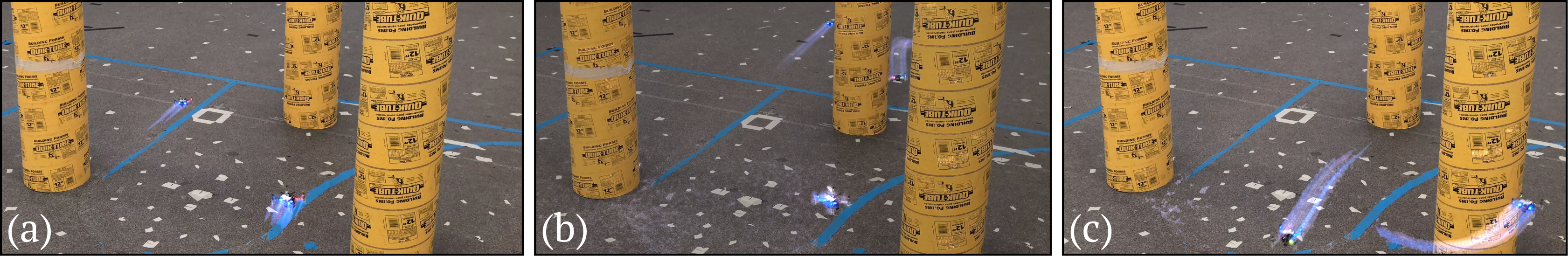}
  \caption{\textbf{High Conflict Negotiation.} The quadrotors are faced with intersecting trajectories with the presence of obstacles. The right outgoing quadrotor waits for the incoming to pass before swerving to the right to dodge.}
  \label{fig:negotiation}
\end{figure*}
\begin{figure}[t!]
\centering
\includegraphics[width=0.48\textwidth]{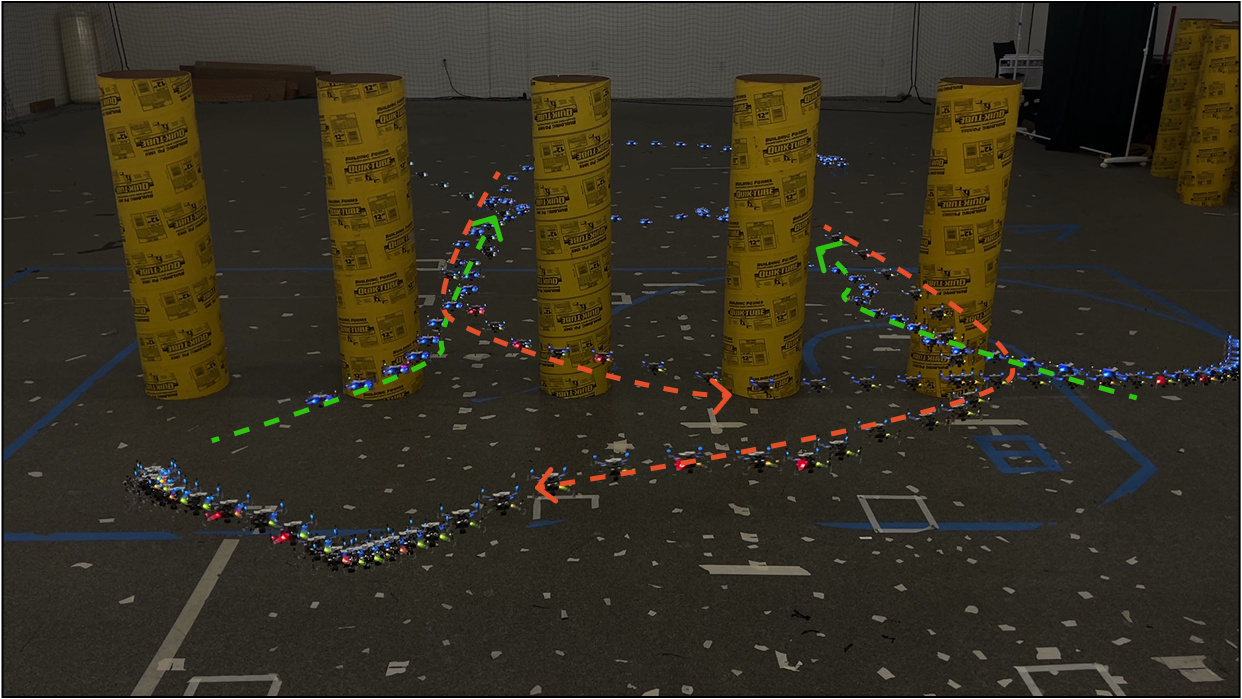}
  \caption{\textbf{Bilateral Swap.} A bilateral swap configuration is shown where quadrotors are tasked with swapping goals along the diagonals. We find that the policy does not travel in the direction of the blind spots, despite the naive trajectory being planned through the diagonals of the area.}
  \label{fig:bilateral_swap}
\end{figure}
To further assess policy deployment, we introduce a bilateral swap scenario, where quadrotors must cross along diagonals of the square resulting in head-on trajectories through the center of the obstacle wall. This is a significantly harder configuration due to tighter interactions in path intersection and reduced clearance near the center. We observe that the policy is still capable of executing the bilateral swap reliably in $0.5m$ gaps. Each quadrotor adjusts its path by slowing down or shifting laterally to avoid collisions through the constrained portion of the space. Notably, these behaviors emerge without explicit coordination logic or trajectory handshaking. Furthermore, the hardware experiments show that the policy learns to only travel in directions within the sensing field of view. This stems from the explicit lateral movements, despite the planned trajectory crossing through the middle. These results reinforce the learned controller’s capability for real-time, decentralized negotiation in high-conflict scenarios even under obstacle restricted environments. We emphasize that the obstacle configuration is unknown \textit{a priori} making the swap more difficult. 
\paragraph{High-Conflict Scenario} 
Finally, Figure \ref{fig:negotiation} shows a challenging real-world deployment scenario involving three quadrotors navigating through a triangular arrangement of obstacles. The quadrotors are initially positioned on intersecting paths, one approaching from the incoming direction, and two departing along outgoing trajectories. At first, the left outgoing quadrotor travels through the obstacle formation (a). The incoming quadrotor must then select a maneuver that allows it to bypass the central obstacle without obstructing the outbound paths of the two others. As it begins its approach, the right outgoing quadrotor preemptively stops tracking the trajectory (b), creating space for the incoming quadrotor. When the incoming has decided to move, the right outgoing then dodges to the right of the obstacle (c). This interaction unfolds without a centralized controller, prior communication or a prebuilt map; instead, each quadrotor operates using only instantaneous state exchanges and local ranging sensing. This scenario highlights our approach’s capability for agile, autonomous conflict resolution in tight, spatially constrained conditions, for multi quadrotor navigation.

%% file: Tables/main_comparison.tex
\begin{table*}[!t]
\centering
\caption{\textbf{Quantitative Comparisons.} Performance comparison across methods, scenarios, and planner versions. All rates are fractions over $N=50$ trials with 8 quadrotors; collision rates, distances, speeds are $\mu \pm \sigma$. Q-Q coll. denotes quadrotor to quadrotor collision fractions, whereas quadrotor to obstacle fractions are denoted by Q-O coll. The swap goal task becomes significantly more difficult due to intersections between quadrotors and obstacles.}
\label{tab:comparison}
\begin{threeparttable}
\resizebox{\textwidth}{!}{%
    \begin{tabular}{@{}llccccccccc@{}}
        \toprule
        \textbf{Method} & \textbf{Scenario} 
        & \textbf{Quadrotor‐level $\uparrow$} 
        & \textbf{Overall\tnote{*} $\uparrow$} 
        & \textbf{Incomplete\tnote{*} $\downarrow$} 
        & \textbf{Q–Q coll. $\downarrow$} 
        & \textbf{Q–O coll. $\downarrow$} 
        & \textbf{Avg. Dist. $\downarrow$} 
        & \textbf{Avg. Vel. $\uparrow$} 
        & \textbf{Avg. Acc. $\downarrow$} 
        & \textbf{Avg. Jerk $\downarrow$} \\
        \midrule
        \textbf{LEARN}
        & \textit{Straight Line} 
        & 0.975  & 0.880 & $0.000 \pm 0.000$  
        & $0.020 \pm 0.069$ & $0.020 \pm 0.058$ 
        & $12.447 \pm 0.226$ & $0.495 \pm 0.007$ 
        & $0.057 \pm 0.011$ & $0.126 \pm 0.054$ \\
        
        & \textit{Swap Goal} 
        & 0.843  & 0.480 & $0.240 \pm  0.431$              
        & $0.100 \pm 0.184$  & $0.083 \pm 0.112$  
        & $13.395 \pm 0.721$ & $0.526 \pm 0.024$ 
        & $0.081 \pm 0.021$ & $0.206 \pm 0.132$ \\
        \addlinespace
        
        ES2 \cite{zhou2022swarm} 
        & \textit{Straight Line} 
        & 0.950 & 0.780 & $0.180 \pm 0.388$  
        & $0.020 \pm 0.035$  & $0.120 \pm 0.075$  
        & $13.689 \pm 1.330$ & $0.924 \pm 0.162$ 
        & $0.063 \pm 0.035$ & $0.205 \pm 0.411$ \\
        
        & \textit{Swap Goal} 
        & 0.860 & 0.440 & $0.280\pm0.453$          
        & $0.042 \pm 0.098$  & $0.074 \pm 0.112$  
        & $13.022 \pm 1.442$ & $0.855 \pm 0.204$  
        & $0.083 \pm 0.046$ & $0.384 \pm 0.500$ \\
          
        \addlinespace
        HDSM \cite{toumieh2024hdsm} 
        & \textit{Straight Line} 
        & 0.765  & 0.300 & $0.160\pm0.370$              
        & $0.215\pm 0.195$  & $0.027\pm 0.068$            
        & $13.022 \pm 0.196$ & $0.414 \pm 0.011$ 
        & $0.057 \pm 0.010$ & $0.271 \pm 0.041$ \\
        
        & \textit{Swap Goal}  
        & 0.735 & 0.340 & $0.040\pm0.198$               
        & $0.271\pm0.249$ & $0.002 \pm 0.016$             
        & $14.921 \pm  1.062$ & $0.127  \pm  0.013$
        & $0.058 \pm 0.011$ & $0.273 \pm 0.044$\\
        
        \bottomrule
    \end{tabular}%
}
\begin{tablenotes}[flushleft]
\footnotesize
\item[*] Denotes that the statistic was calculated across runs, whereas others were calculated across quadrotors.
\end{tablenotes}
\end{threeparttable}
\vspace{-0.5em}
\end{table*}

%% file: Tables/comm_freq.tex
\begin{table}
\centering
\caption{\textbf{Effects of Communication Delays.} The effects of slower state exchange rates on the RL policy for the straight and swap scenarios ($N=50$). All experiments use 8 quadrotors.}
\label{table:comm}
\resizebox{.48\textwidth}{!}{
  \begin{tabular}{@{}llcccc@{}}
  \toprule
  Scenario & \textbf{Frequency ($hz$ / $ms$)}
      & \textbf{Quadrotor‐level} 
      & \textbf{Overall} 
      & \textbf{Q–Q coll.} 
      & \textbf{Q–O coll.} \\
    \midrule
  Straight & 50 $\approx$ 20 & 0.975 & 0.880 & 0.080 & 0.120  \\
  & 45 $\approx$ 22.222             & 0.974 & 0.878 & 0.015 & 0.025 \\
  & 35 $\approx$ 28.571        & 0.966 & 0.837 & 0.020 & 0.017 \\
  & 25 $\approx$ 40            & 0.961 & 0.837 & 0.031 & 0.028\\
  & 15 $\approx$ 66.667             & 0.974 & 0.857 & 0.015 & 0.012\\  
  & 5 $\approx$ 200     & 0.977 & 0.878 & 0.010 & 0.020 \\
  \midrule
  Swap & 50 $\approx$ 20            & 0.864           &  0.510 & 0.089 & 0.081 \\
  & 45 $\approx$ 22.222             & 0.859           & 0.408 &  0.076 & 0.092 \\
  & 35 $\approx$ 28.571            & 0.877           & 0.489 & 0.089 & 0.079\\
  & 25 $\approx$ 40            & 0.877    & 0.530 & 0.066 & 0.089 \\
  & 15 $\approx$ 66.667             & 0.869    & 0.510 & 0.0841 & 0.076 \\
  & 5 $\approx$ 200        & 0.885    & 0.489 & 0.061 & 0.059 \\
  \bottomrule
  \end{tabular}
}
\end{table}

%% file: Sections/Conclusion.tex
\section{Limitations}
While our results demonstrate the feasibility of deploying a fully onboard learned navigation policy for multi-robot coordination, several limitations remain.

First, the policy relies on local sensing and short-horizon reasoning, which can lead to suboptimal behavior in environments requiring long-term foresight—such as dead-ends, maze-like corridors, or traps requiring backtracking. Without global map information or memory mechanisms, the system must rely on reactive adjustments, which may not suffice in adversarial or partially observable settings. As our method works at the control level, future work can extend by integrating a relatively lightweight motion planner, where the neural controller is trained to correct for potential mistakes. This would integrate both a long horizon decision making mechanism with our short horizon lightweight observation conditioned controller. 

Second, while the system is trained to generalize across a range of obstacle configurations, it assumes a relatively static environment during deployment. Rapidly moving obstacles, strong wind disturbances, or unpredictable third-party agents (e.g., humans or other robots) may challenge the policy's reactivity. One can include disturbances and faster moving objects within training to improve the controller. 

Addressing these limitations will involve integrating richer sensor fusion, predictive modeling, memory-aware policies, and adaptive communication strategies—opening pathways for more robust and scalable decentralized multi-robot navigation. However, given the current limits of the platform we believe that our method pushes the hardware constraints to its limit.
\section{Conclusion}
We present an end-to-end learned navigation system designed for real-time deployment on resource-constrained nano quadrotors. Our approach integrates a lightweight attention-based neural network with minimal depth sensing, enabling fully onboard perception, planning, and control without reliance on external infrastructure or centralized coordination.

Through extensive simulation and hardware experiments, we demonstrated that our policy can reliably perform multi-quadrotor navigation, obstacle avoidance, and formation tracking in both unknown environments. The system maintains low collision rates even in dense scenarios, and generalizes across a wide range of obstacle configurations and quadrotor formations. Our results indicate marginally better performance over state of the art motion planners, despite only using a fraction of compute. Notably, we showed that the learned controller can handle high-conflict maneuvers including goal-swapping, bilateral negotiations, and conflict resolution without explicit communication or global planning. These results underscore the feasibility of using compact, learned attention models for decentralized multi-quadrotor navigation in highly constrained settings. Future work will explore scaling to larger teams, incorporating more diverse sensor modalities, and integrating memory for long-horizon reasoning in dynamic environments.